\documentclass[10pt,twocolumn,letterpaper]{article}

\usepackage{arxiv}
\usepackage{times}
\usepackage{epsfig}
\usepackage{graphicx}
\usepackage{amsmath}
\usepackage{amssymb}
\usepackage[pagebackref=true,breaklinks=true,colorlinks,bookmarks=false]{hyperref}
\usepackage{booktabs}
\usepackage{placeins}
\usepackage{multirow}
\usepackage[accsupp]{axessibility}
\usepackage{xcolor}

\arxivfinalcopy

\ifarxivfinal\fi
\begin{document}

\title{Ev-NeRF: Event Based Neural Radiance Field}

\author{Inwoo Hwang$^1$, Junho Kim$^1$, and Young Min Kim$^{1, 2,}$\thanks{Young Min Kim is the corresponding author.}
\\
\small
$^1$Department of Electrical and Computer Engineering, Seoul National University\\
\small
$^2$Interdisciplinary Program in Artificial Intelligence and INMC, Seoul National University\\
}

\maketitle
\begin{abstract}
We present Ev-NeRF, a Neural Radiance Field derived from event data.
While event cameras can measure subtle brightness changes in high frame rates, the measurements in low lighting or extreme motion suffer from significant domain discrepancy with complex noise.
As a result, the performance of event-based vision tasks does not transfer to challenging environments, where the event cameras are expected to thrive over normal cameras.
We find that the multi-view consistency of NeRF provides a powerful self-supervision signal for eliminating spurious measurements and extracting the consistent underlying structure despite highly noisy input.
Instead of posed images of the original NeRF, the input to Ev-NeRF is the event measurements accompanied by the movements of the sensors.
Using the loss function that reflects the measurement model of the sensor, Ev-NeRF creates an integrated neural volume that summarizes the unstructured and sparse data points captured for about 2-4 seconds.
The generated neural volume can also produce intensity images from novel views with reasonable depth estimates, which can serve as a high-quality input to various vision-based tasks.
Our results show that Ev-NeRF achieves competitive performance for intensity image reconstruction under extreme noise and high-dynamic-range imaging.

\end{abstract}

\section{Introduction}
\label{sec:intro}

Event cameras are neuromorphic sensors, where individual pixels detect changes of brightness that exceed a threshold.
The output of event cameras is a sequence of asynchronous events composed of the polarity, pixel location, and the time stamp, occurring only at a sparse set of locations where the brightness change is detected.
They have many advantages over conventional cameras such as high temporal resolution, low energy consumption, and high dynamic range \cite{Gallego:2020:survey}.
However, the measurements of the same object change significantly under different motion or lighting conditions causing domain discrepancy in real-world deployment~\cite{Kim_2021_ICCV,amae,denoise_exp_special}.
While event cameras are expected to prosper under extreme environmental conditions, the performance of event-based vision tasks often deteriorates due to the significant domain shift with severe noise.

\begin{figure*}[ht!]
    \centering
    \includegraphics[width=\linewidth]{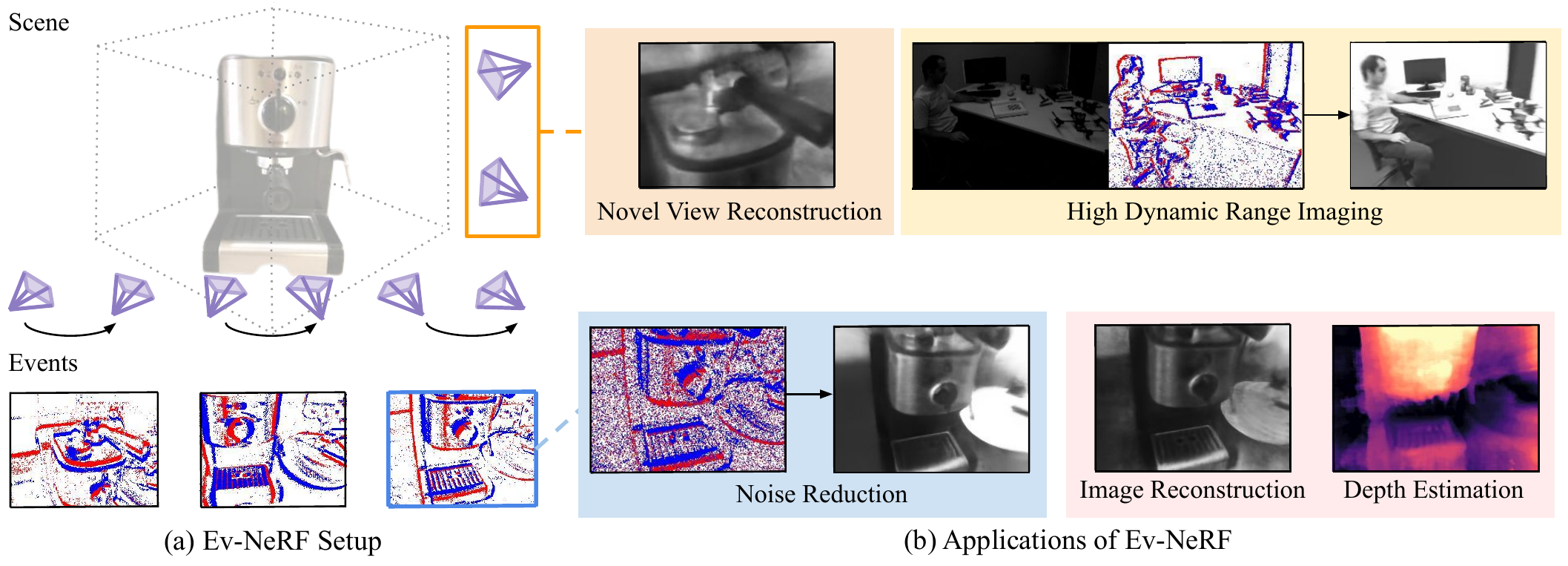}
    \caption{(a) Ev-NeRF operates with event data obtained from a moving event camera. (b) Ev-NeRF learns the implicit volume with the raw event output of the sensor and serves as a solution for various event-based applications, such as high dynamic range imaging, noise reduction, depth estimation, intensity image reconstruction, and novel-view intensity image reconstruction.}
\label{fig:teaser}
\end{figure*}

The output of event streams is very different from an ordinary image, which is a two-dimensional array with dense color values.
Many existing approaches using event data compile them into a more structured form for denoising~\cite{Czech:2016:denoise-spatial,DelbruckL:2008:denoise-spatial,Liu:2015:denoise-spatial-temporal,Feng:2020:denoise-spatialtemporaldensity,baldwin:2020:denoise,Duan:2021:denoise}, or directly perform downstream tasks such as motion estimation \cite{Mitrokhin:2020:motion,liu:2020:motion,Nunes:2020:motion,Stoffregen:2019:motion} or  pose estimation \cite{Mueggler:2017:dataset,Bryner:2019:pose}.
Nonetheless, training data is often limited and the performance of event-based vision is often inferior to the performance of the same tasks with conventional images~\cite{Kim_2021_ICCV}.
The complex noise characteristics and domain discrepancy further complicate developing practical algorithms for event cameras.

Inspired by the recent success of Neural Radiance Fields (NeRF) \cite{mildenhall:2020:nerf}, we propose Ev-NeRF, a neural radiance field built directly from raw event data, as shown in Figure~\ref{fig:teaser}(a).
Ev-NeRF builds a 3D volumetric representation that can concurrently explain events associated with the camera movement.
Given the 5D input of location and viewing direction, NeRF outputs the volume density and emitted color, which can be aggregated to synthesize an image from an arbitrary viewpoint by the volume rendering.
While NeRF is trained to minimize the color discrepancy between the synthesized image and the ground truth image, Ev-NeRF is trained with a new loss function that incorporates the sensor movement and the resulting events triggered by the difference of brightness.

Ev-NeRF properly handles the complex noise in event cameras without ground truth supervision, and at the same time, enjoys the technical advantages of the sensor over conventional cameras.
The volumetric aggregation in the formulation effectively reduces the prevalent noise in event measurements \cite{Gallego:2020:survey}, as the complex spatial and temporal noises lack multi-view consistency.
Further, the associated intensity values in Ev-NeRF are in a high dynamic range (HDR), as the provided measurements of event cameras are sensitive to extreme lighting beyond the dynamic range of the conventional camera.

Interestingly, the created volumetric representation is a solution to many of the vision problems tackled in previous works using event data, as shown in Figure~\ref{fig:teaser}(b).
While the event data only contain the relative changes in the brightness instead of the absolute term, the trained volume can synthesize the intensity image for ordinary computer vision, which is one of frequently tackled problems in the community \cite{Cook:2011:intensity,Kim:2014:intensity,Bardow:2016:intensity,Rebecq:2019tpami:intensity,Rebecq:2019cvpr:intensity,Rebecq:2018:simulator,paredes:2021:SSL-intensity-baseline,Stoffregen:2020:SOTA,Weng:2021:intensity,Scheerlinck:2020:intensity,Wang:2019:HDR,cadena:2021:intensity,wang:2020:intensity}.
Further, the reconstructed density volume can represent the approximate 3D structure of the scene.
This is inherent from the original NeRF formulation enforcing the multi-view consistency, and the quality of 3D reconstruction is superior to the 3D structure built from previous approaches \cite{Kim:2016:3d,Rebecq:18:3d,zhou:2018:3d}.

Our contributions can be summarized as follows: 
\begin{itemize}
    \item We suggest Ev-NeRF, which combines the popular NeRF formulation with the raw event output of a neuromorphic camera for the first time.
    \item Ev-NeRF is highly robust to event noise and builds a coherent 3D structure that can provide high-quality observations.
    \item The created neural volume serves as solutions for various event-based applications, namely intensity image reconstruction,  novel-view image synthesis, 3D reconstruction, and HDR imaging.
    \item Ev-NeRF demonstrates performance comparable to many of existing event-vision algorithms that are dedicated to a specific task in the experimental result.
\end{itemize}
Given the strong experimental result, we expect Ev-NeRF to expand the possible application area of event-based vision that fully leverages the potential of the sensor.

\section{Related works}
\label{sec:related}

In this section, we review the key tasks in event-based vision, along with existing work on neural implicit 3D representations.

\paragraph{Processing Event Data}
Although event cameras can acquire visual information in challenging conditions such as low-lighting or extreme motion, a significant domain gap occurs due to the large amount of noise which further leads to performance degradation~\cite{Kim_2021_ICCV,hats,megapixel,amae,ev-tta}.
Wu et al.~\cite{denoise_exp_special} first demonstrated that event-based vision can deteriorate due to increased noise levels, although the assessments were mainly conducted in synthetic events.
Kim et al.~\cite{Kim_2021_ICCV} further introduced a large-scale dataset enabling systematic assessment of object recognition tasks, and demonstrated that large camera motion or illumination change leads to greater amounts of noise that ultimately deteriorates performance.
Existing approaches denoise the raw data to cope with such adversaries~\cite{denoise_exp_special,ev_gait,noise1}, or suggest stacking events to overcome domain gaps under extreme lighting condition~\cite{hdr1}.
On the other hand, Ev-NeRF can compensate for the spurious noises by enforcing multi-view consistency for the scene geometry.

Instead of handling complex data characteristics from the raw data, many approaches aggregate the sequential measurements into an ordinary image or 3D geometry.
Early attempts for intensity image reconstruction are inspired from statistical methods~\cite{Cook:2011:intensity,Kim:2014:intensity,Bardow:2016:intensity}.
Several subsequent approaches suggest various network architecture designs to improve the image quality or computational cost~\cite{Scheerlinck:2020:intensity,Wang:2019:HDR,cadena:2021:intensity,wang:2020:intensity,Weng:2021:intensity}.
Because the sensor has a high dynamic range, the intensity image restoration can be explicitly designed for HDR images~\cite{Zou:2021:HDR,Wang:2019:HDR} or by applying domain adaptation to day-light condition~\cite{Zhang:2020:HDR_domainadaptation}.
For estimating 3D geometry, recent event-based SLAM methods utilize classical techniques~\cite{Kim:2016:3d,Kim:2014:intensity,ev_slam_2,ev_slam_3,zhou:2018:3d,ev_slam4}, minimizing the energy function formulated over the image-like event representations.
On the other hand, for event-based depth estimation both classical method~\cite{depth_3,depth_2} and learning-based approaches~\cite{Hidalgo:2020:depth,depth_1} coexist.

However, for any of the aforementioned tasks, it is challenging to obtain a large-scale dataset with the ground-truth label or formulate the correct measurement model that represents the wide range of possible sensor characteristics.
~\cite{Rebecq:2019cvpr:intensity,Rebecq:2019tpami:intensity} suggested generating the training data using simulator~\cite{Rebecq:2018:simulator}.
\cite{Stoffregen:2020:SOTA} examined the statistical aspect to reduce the gap.
\cite{paredes:2021:SSL-intensity-baseline} proposed a self-supervised learning framework with the aid of optical flow and does not require ground truth, but their reconstructed images are characterized by several artifacts.
On the contrary, Ev-NeRF works without the ground truth or synthetic data and shows stable results comparable to the state-of-the-art in intensity image reconstruction or depth estimation.

\paragraph{Neural Implicit 3D Representation}
Neural Implicit 3D Representation is gaining popularity due to its strong advantage of memory requirements, no restrictions on spatial resolution, and representation capability. 
Several works~\cite{Park:2019:deepsdf,Mescheder:2019:OccupancyNetworks,Chen:2019:IMnet} showed the advantage of neural implicit representations with 3D supervision.
NeRF (Neural Radiance Fields)~\cite{mildenhall:2020:nerf} proposes an implicit representation of 3D coordinates and viewing direction which can synthesize images with volume rendering techniques.
The resulting neural volume contains information about 3D volume density and emitted radiance for rendering images.
Motivated by the photo-realistic quality of the produced images, a large number of subsequent works spurred to overcome the limitation of the original NeRF including: enabling fast convergence and rendering~\cite{mueller2022instant,sun2021direct,yu2021plenoxels}; handling the input images with unknown or noisy camera poses~\cite{wang2021nerfmm,lin2021barf}; recovering hdr scenes with noisy raw images~\cite{mildenhall2021rawnerf}; or processing dynamic scenes~\cite{dynamic1,dynamic2}.
Our method learns the NeRF volume with event data.
By enforcing the multi-view consistency of collected measurements, Ev-NeRF produces a high-quality image or depth in a novel view and effectively removes spurious noises of an event camera.
While there also exists a concurrent work learning implicit volume with events~\cite{rudnev2022eventnerf}, Ev-NeRF extensively reveals the practical capability.

\begin{figure*}[t!]
    \centering
    \includegraphics[width=\linewidth]{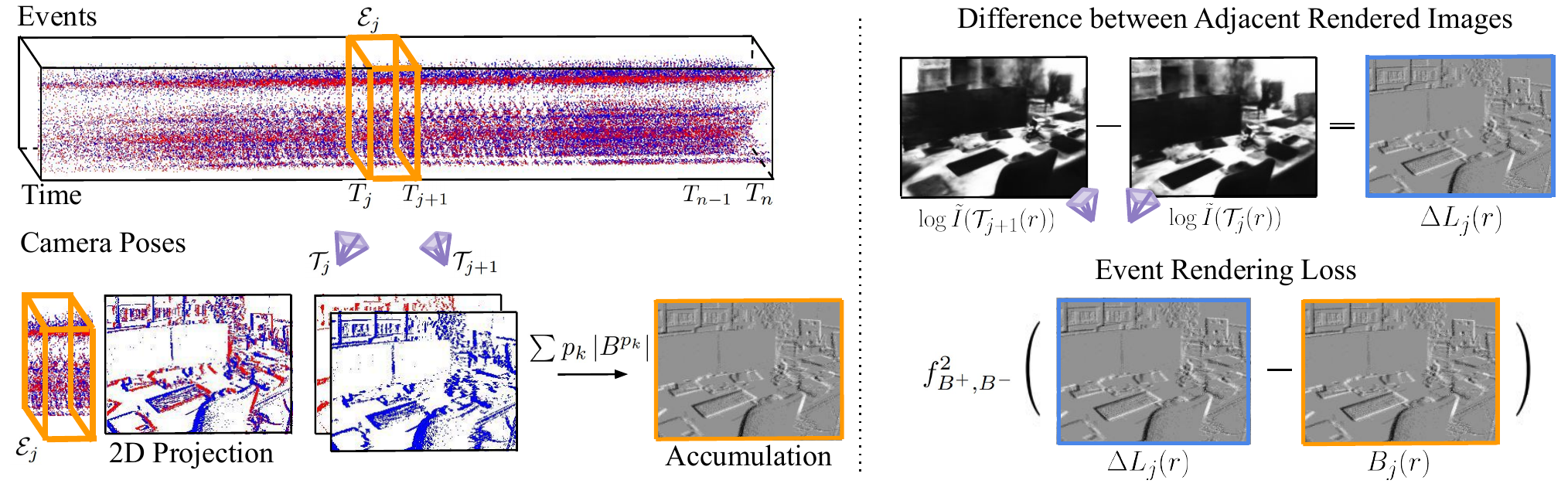}
    \caption{Overview of our method. According to the measurement model of the sensor, the events $\mathcal{E}_j$ accumulated during a short time interval $[T_j, T_{j+1})$ should reflect the difference in brightness.
    Using the implicit volume, we render intensity frames from the view points of two adjacent event camera poses, $\mathcal{T}_j$ and $\mathcal{T}_{j+1}$. 
    Event rendering loss is the discrepancy between the accumulated event $B_j(r)$ and difference in the intensity of adjacent rendered frames $\Delta L_j(r)$.}
\label{fig:event_rendering_loss}
\vspace{-0.9em}
\end{figure*}

\section{Background}
\label{sec:background}

For the completeness of discussion, we include the event generation model of the sensor followed by the mathematical formulations of the neural radiance fields (NeRF), which serves as the two main components for deriving Ev-NeRF.

\paragraph{Event Generation Model}
Instead of recording the absolute color values of the image pixels, an event camera records asynchronous changes of the brightness as a sequence of events $E_{k}=\left( u_{k},v_{k},t_{k},p_{k}\right)$, indicating that the brightness change at the pixel coordinate $\left( u_{k},v_{k}\right)$ reaches a specific threshold $B$ at time $t_{k}$,
\begin{equation}
|L\left(u_k, v_k,t_k\right) -L(u_k, v_k ,t_k-\delta t)| \geq \left|B\right|,
\label{eq:delta_L}
\end{equation}
where $L=\log \left( I\right)$ is the logarithm of brightness $I$ and $\delta t$ is time that has passed since the last event.
$p_k \in \{+,-\}$ is the polarity denoting whether the brightness change is positive or negative.
It is known that the threshold for triggering positive events is different from the one for the negative events~\cite{Gallego:2020:survey}, which we denote $B^{+}$ and $B^{-}$, respectively.
If we accumulate the events occurring for a given period of time $\Delta t$, the brightness change in a specific pixel can be approximated by~\cite{Gallego:2020:survey}
\begin{equation}
\Delta L\left(u, v,\Delta t\right) =\sum_{\substack{t_{k}\in \Delta t, \\(u_k,v_k)=(u,v)}} p_{k}\left|B^{p_{k}}\right|.
\label{eq:delta_L_sum}
\end{equation}
The threshold $B^{p_{i}}$ can be different under various physical conditions, which further challenges the event-based vision, in addition to complex noise characteristics of the sensor.

\paragraph{Neural Radiance Fields}
Ev-NeRF takes inspiration from NeRF~\cite{mildenhall:2020:nerf} which is trained to accumulate the volumetric information with 2D supervision.
The supervision signal for NeRF is the total squared error between the rendered and true pixel colors.
Basically the neural network $F_\theta(\cdot)$ receives the input of the 3D coordinate $\mathbf{x}_i \in \mathbb{R}^3$ and the ray direction $\mathbf{d}_i \in \mathbb{S}^2$ and outputs the density $\sigma_{i} \in \mathbb{R}$ and emitted radiance $\textbf c_{i} \in \mathbb{R}^3$
\begin{equation}
F_\theta: (\gamma_x(\mathbf{x}_i), \gamma_d(\mathbf{d}_i)) 
\rightarrow (\sigma_i, \mathbf{c}_i).
    \label{eq:nerf}
\end{equation}
Here $\gamma \left( \cdot \right)$ is sinusoidal positional encoding function which successfully captures  the high-frequency information along the spatial direction.
With positional encoding and coarse-to-fine sampling techniques, a neural network is trained to synthesize high-quality novel view images.

Following the classical volume rendering technique, each pixel is rendered by sampling $N$ points $\mathbf{x}_{1},\ldots ,\mathbf{x}_{N}$ of the volume density along the ray $r(\mathbf{x}_0, \mathbf{d})$. 
$\mathbf{x}_0$ is the initial point of the ray located at the focal point of the camera using the pinhole camera model.
The final rendered color of the pixel is aggregated along the ray as
\begin{equation}
\hat{C}\left( r\right) =\sum^{N}_{i=1}A_{i}\alpha _{i} \textbf c_{i}.
\label{eq:nerf_color}
\end{equation}
$A_{i}=\exp \left( -\sum ^{i-1}_{l=1}\sigma_{l}\delta _{l}\right)$ denotes the accumulated transmittance along the ray, and 
$\alpha_{i}=1-\exp \left( -\sigma_{i}\delta _{i}\right)$ denotes the alpha value, where $\delta_{i}=\left\| \mathbf{x}_{i+1}-\mathbf{x}_{i}\right\|$
is the distance between adjacent samples.
Additionally, the depth along the ray direction can be approximated with a similar formulation:
\begin{equation}
\hat{D}\left( r\right) =\sum ^{N}_{i=1}A_{i}\alpha _{i} s_{i},
\label{eq:nerf_depth}
\end{equation}
where $s_{i}$ denotes the distance between $\mathbf{x}_0$ and $\textbf{x}_{i}$.

\section{Method}

Ev-NeRF creates neural implicit representation $F_\theta$ of a static scene as NeRF.
Since an event is triggered when the brightness changes, we use a slice of the event sequence $\mathcal{E}_j=\{E_k=(u_k, v_k, p_k, t_k) | T_j \leq t_k < T_{j+1}\}$ during a  small duration of time $[{T}_j, {T}_{j+1})$.
The motion of the camera is provided by the starting and ending poses of the period, $\mathcal{T}_j$ and $\mathcal{T}_{j+1}$.
The neural network for Ev-NeRF regresses for one dimensional emitted luminance value $y_i \in \mathbb{R}$ instead of RGB color values,
\begin{equation}
F_{\theta }:\left( \gamma_{x}\left(\mathbf{x}_i\right) ,\gamma_{d}\left(\mathbf{d}_i\right) \right) \rightarrow \left( \sigma_i ,y_i\right).
\label{eq:ev_nerf}
\end{equation}
This is a natural choice considering that the event only records the brightness change in a single channel.

After the neural network is trained, we can render the intensity image from an arbitrary viewpoint adapting Equation~\ref{eq:nerf_color}.
To elaborate, if we define the camera ray $r$ that passes through the pixel location $(u, v)$ from a camera pose $\mathcal{T}$, we can sample $N$ points along the ray and apply volume rendering technique to find the intensity of the pixel
\begin{equation}
\hat{I}(\mathcal{T}(r))=\sum_{i=1}^N A_i \alpha_i y_i.
    \label{eq:ev_nerf_intensity}
\end{equation}
The depth measurement can also be approximated using Equation~\ref{eq:nerf_depth}.

When we combine the formulation with the event generation model, we also jointly optimize for the unknown thresholds, $B^+_j$ and $B^-_j$, in addition to the implicit neural volume $F_\theta(\cdot)$.
We assume that the threshold is a function of time and polarity, but is spatially the same for all pixels.
More specifically, we assume that the threshold is constant in each time interval $[{T}_j, {T}_{j+1})$, but changes when the time interval changes.

The total loss used to train Ev-NeRF is given by
\begin{equation}
\mathcal{L}_\text{total} = \mathcal{L}_\text{event} + \lambda\mathcal{L}_\text{thres}.
\label{eq:total_loss}
\end{equation}
$\mathcal{L}_\text{event}$ is the event rendering loss  which replaces the image rendering loss in conventional NeRF. 
Here we combine the event generation model in Section~\ref{sec:background} with the volume rendering formulation of the original NeRF.
$\mathcal{L}_\text{thres}$ is the threshold bound loss,  designed to avoid degenerate cases.

\paragraph{Event Rendering Loss}
\label{sec:loss_ev}
Our loss for training $F_\theta$ compares the recorded events and the difference in the rendered intensity, as shown in Figure~\ref{fig:event_rendering_loss}.
Let us denote the intensity values of the pixel ray $r$ at time ${T}_j$ and ${T}_{j+1}$ as  $\hat{I}(\mathcal{T}_{j}(r))$ and $\hat{I}(\mathcal{T}_{j+1}(r))$, respectively, where the intensity images are obtained using Equation~\ref{eq:ev_nerf_intensity}.
Then we can calculate $\Delta L_j$ at the pixel ray $r$ as
\begin{equation}
\Delta L_{j}(r) = \log \hat{I}\left( \mathcal{T}_{j+1} (r)\right) - \log \hat{I}\left( \mathcal{T}_{j} (r)\right).
\label{eq:delta_L_ev_nerf}
\end{equation}
Using the event generation model in Equation~\ref{eq:delta_L_sum}, $\Delta L_j(r)$ in Equation~\ref{eq:delta_L_ev_nerf} should be measured by the accumulated sum of the events $B_j(r) = \sum_{E_k \in \mathcal{E}_j(r)} p_k B_j^{p_k}$ within the time interval $[{T}_j, {T}_{j+1})$.

Our event-rendering loss $\mathcal{L}_\text{event}$ is the total sum of discrepancy for all time intervals $j$ and rays $r$ available in the batch,
\begin{equation}
\mathcal{L}_\text{event} = \sum_j\sum_{r} f_{B_j^{+},B_j^{-}}^{2} (\Delta L_j (r) - B_j(r)),
\label{eq:loss_event}
\end{equation}
where the function $f$ penalizes the discrepancy above the sensor threshold by incorporating a dead zone $\left[B^{-}, B^{+}\right]$ as described in~\cite{Bardow:2016:intensity}:
\begin{equation}
f_{B^{+},B^{-}} (x) = 
\begin{cases}
x - B^{+}, & \mbox{if } x > B^{+},
\\
0, & \mbox{if } B^{-} \leq  x \leq  B^{+},
\\
 - x + B^{-}, & \mbox{if } x < B^{-}.
\end{cases}
    \label{eq:ev_nerf_loss_f}
\end{equation}
Therefore we only focus on where the measurements provide enough evidence for the brightness changes.

\paragraph{Threshold Bound Loss}
\label{sec:loss_th}

While the joint optimization over the unknown threshold values $B_j^+, B_j^-$ improves the performance of Ev-NeRF, the additional  parameters further challenge the optimization process which already is highly under-constrained with the unknown brightness values $I$.
Without additional constraints, we empirically observed that the network often converges to the trivial solution with $\Delta I =0$ and the threshold value $0$.
The threshold bound loss is a simple prior to keep the threshold values within the reasonable bound:
\begin{equation}
\mathcal{L}_\text{thres} = r\left(B_0^{+}- B_{j}^{+}\right) + r\left(B_{j}^{-}-B_0^-\right),
\label{eq:loss_thres}
\end{equation}
where $r\left( \cdot \right)$ is a ReLU function. 
In our experiments, we set $B_0^{+}=0.3$ and $B_0^{-}=-0.3$, based on our prior knowledge about threshold scale \cite{Gallego:2020:survey}.

\begin{figure*}[t!]
\centering
\includegraphics[width=0.95\linewidth]{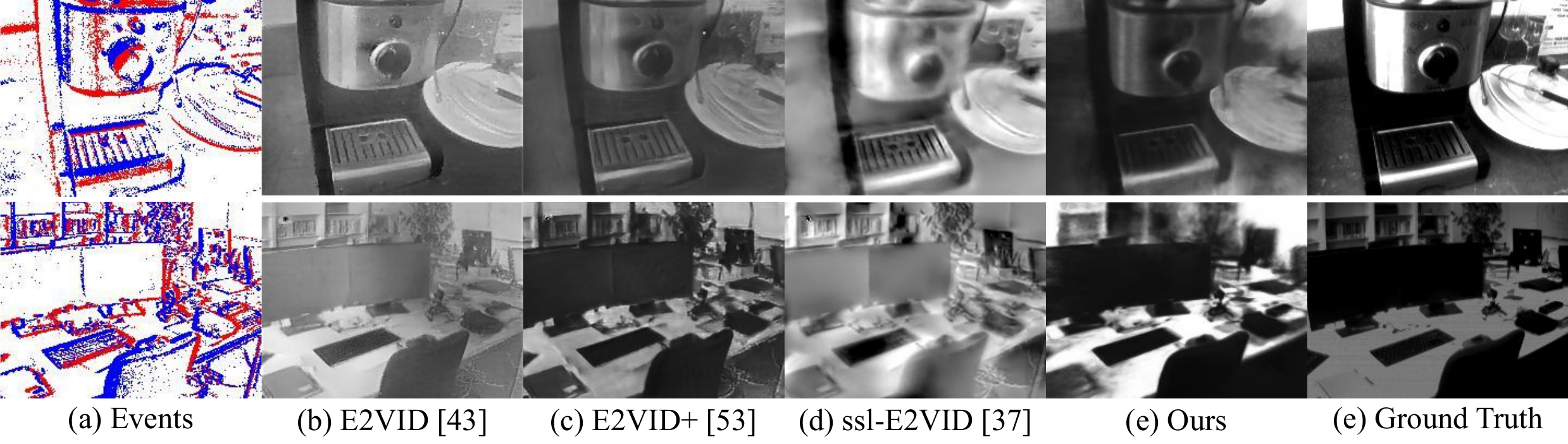}
\caption{Qualitative comparison on intensity image reconstruction.}
\label{fig:intensity_recon}
\vspace{-1.2em}
\end{figure*}

\section{Results}

Once we train Ev-NeRF, from highly corrupted event sequence data, the coherent implicit volumetric representation of NeRF is learned.
The integrated representation can generate images or depth estimates in a novel view, which are compared against previous works or designed baseline.
We evaluate Ev-NeRF on both synthetic and real dataset.

\paragraph{Implementation Detail}
The input to Ev-NeRF is the stream of event data obtained from an event sensor moving around a static scene.
The stream data is accompanied by a sequence of the sensor's intermediate positions which are time-stamped.
The sensor positions can be acquired from an additional sensor or structure from motion (SfM) using intensity images or provided from simulator.

For real world data, we calculate the poses by running SfM~\cite{schoenberger:2016:colmap-sfm,schoenberger:2016:colmap-mvs} with the intensity frames, which are provided in the datasets~\cite{Mueggler:2017:dataset,Stoffregen:2020:SOTA,zhou:2018:3d} recorded at about 24 Hz.
Except for this process, the intensity frames are not available to Ev-NeRF during training and are used only for evaluation.
We use about 50 to 100 consecutive event slices $\mathcal{E}_j$ to train a neural volume, where the length of each time slice $[{T}_j, {T}_{j+1})$ is chosen to be the frame rate of intensity frames, about 1/24s.
The duration of the total event sequence used for Ev-NeRF is roughly 2-4 seconds.
For each event slice $\mathcal{E}_j$ corresponding to $[{T}_j, {T}_{j+1})$, we add random events equivalent to 5\% of the number of events that occurred at time slice $[{T}_j, {T}_{j+1})$ during training.
We find that the additional random noise slightly improves the quality of neural representation in ambiguous regions, which is further described in the supplementary material.

\subsection{Robust Intensity Image Reconstruction}
\label{sec:result_robust}

Ev-NeRF creates the NeRF volume that aggregates multiple observations, which is robust against a variety of perturbation such as the amount of noise or extreme light conditions.
We compare the quality of the synthesized image against previous works on intensity image reconstruction under noisy or low-light condition.
We use three available real-world datasets that are widely used for event-based image reconstruction, namely IJRR~\cite{Mueggler:2017:dataset},  HQF~\cite{Stoffregen:2020:SOTA}, and Stereo DAVIS dataset~\cite{zhou:2018:3d} for quantitative comparison.
For the stereo DAVIS dataset, we only use the measurements from a single event camera.
Unless otherwise noted, we use four sub sequences ({dynamic\_6dof, office\_spiral, office\_zigzag, hdr\_boxes}) from the IJRR dataset, three sequences ({reflective\_materials, high\_texture\_plants, still\_life}) from HQF dataset and two sequences ({monitor, reader}) from Stereo DAVIS dataset.

\paragraph{Intensity Image Reconstruction}
Figure~\ref{fig:intensity_recon} shows exemplar reconstructed images, and different approaches exhibit different kinds of artifacts.
We compare the performance of Ev-NeRF against three baselines: E2VID~\cite{Rebecq:2019cvpr:intensity}, E2VID+~\cite{Stoffregen:2020:SOTA} and ssl-E2VID~\cite{paredes:2021:SSL-intensity-baseline}.
E2VID~\cite{Rebecq:2019cvpr:intensity} is trained in a supervised fashion with a large amount of synthetic data.
E2VID+~\cite{Stoffregen:2020:SOTA} adjusts the synthetic training data to better fit the distribution of the real data.
ssl-E2VID~\cite{paredes:2021:SSL-intensity-baseline} tries to overcome the domain gap and suggests a self-supervised approach achieving results comparable to E2VID+~\cite{Stoffregen:2020:SOTA} without ground truth data.
Unlike Ev-NeRF, which is trained for each scene, previous works are trained in a supervised fashion with a synthetic dataset composed of a pair of ground truth images and event measurements.
A quantitative comparison of image reconstruction on real-world scenes is in the supplementary material.
More importantly, Ev-NeRF only enforces multi-view consistency without further domain-dependent training, whereas previous approaches are trained in a fixed resolution dedicated to a specific measurement condition.
We further validate intensity image reconstruction results on the CED Dataset~\cite{colordataset} with a different resolution (346$\times$240) in the supplementary.

\paragraph{Noise Resistant Image Reconstruction}

\begin{figure}[h!]
\centering
\includegraphics[width=0.9\linewidth]{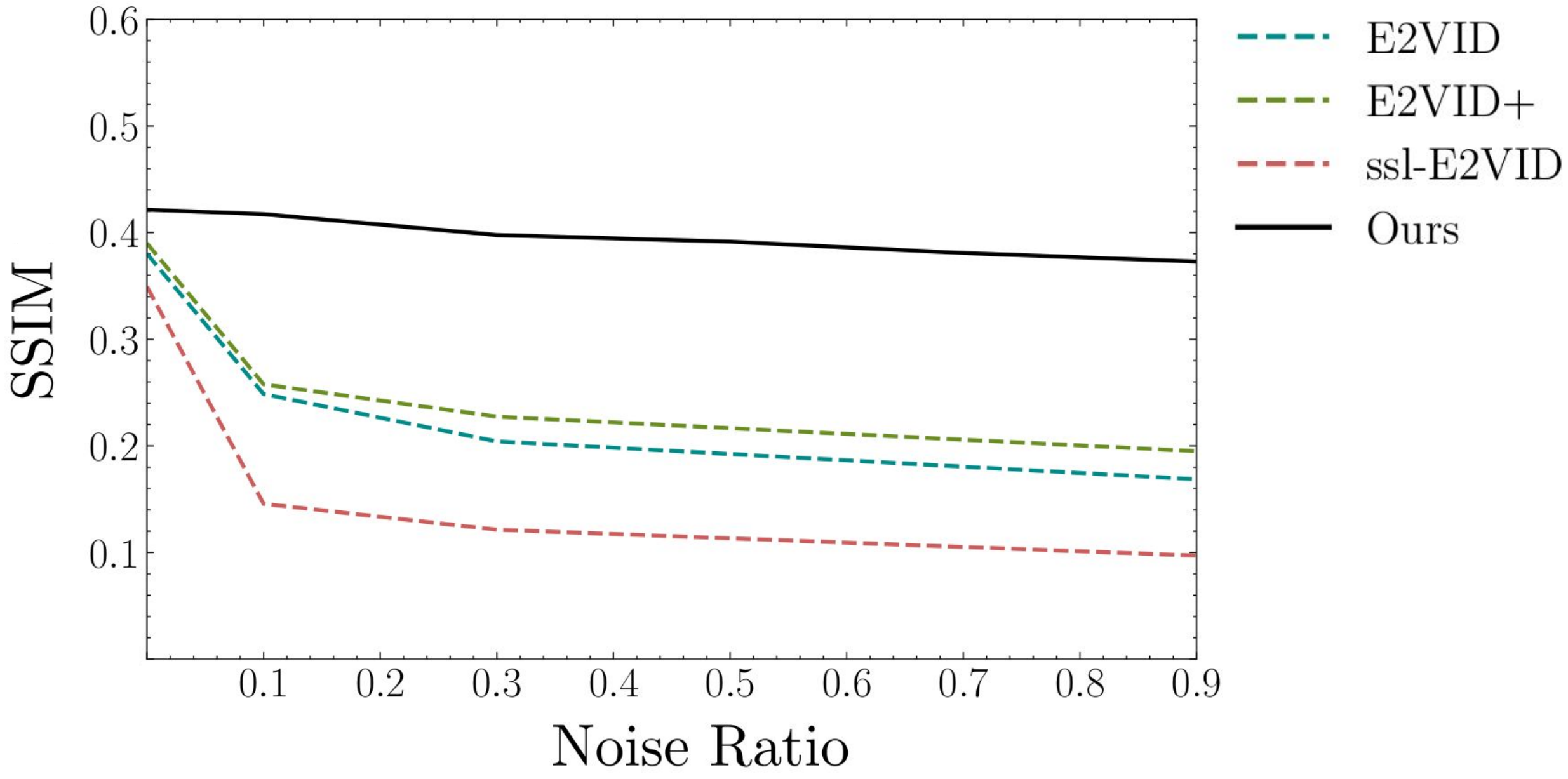}
\caption{Effect of noise on various image reconstruction methods. Even in serious noise conditions, Ev-NeRF can robustly reconstruct images whose quality is comparable to that of other approaches in normal conditions.}
\label{fig:noise_reduction_graph}
\end{figure}

\begin{figure}[ht]
\centering
\includegraphics[width=0.9\linewidth]{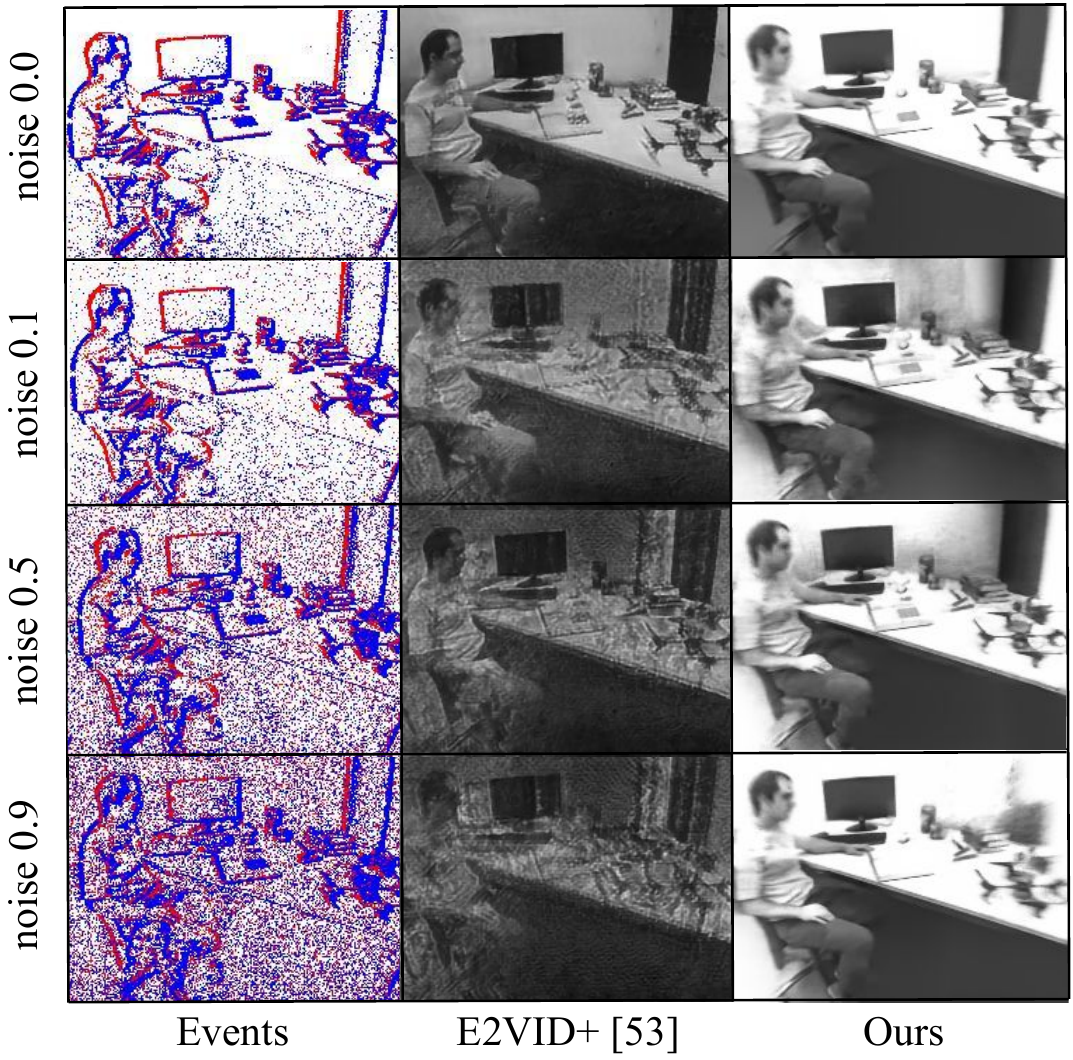}
\caption{
Qualitative comparison on image reconstruction given input with various noise levels. 
Ev-NeRF shows little performance degradation even with extremely noisy inputs.}
\label{fig:noise_reduction}
\vspace{-0.9em}
\end{figure}

The performance gap becomes prominent when the measurements exhibit severe noises, introducing domain shift.
We use the event camera simulator v2e~\cite{v2e} to synthetically add realistic sensor noise due to photon fluctuations or invalid threshold values.
The simulator allows us to control the amount of noise, which we indicate with the ratio of the number of noisy events added to the number of existing events.
Figure~\ref{fig:noise_reduction_graph} evaluates the SSIM of the reconstructed intensity image compared against the ground truth intensity image with the office\_zigzag scene in the IJRR~\cite{Mueggler:2017:dataset}.
The effectiveness of Ev-NeRF is prominent for noisy events, where it compensates the complex noise despite over 70\% of noise and achieves comparable quality as the state-of-the-art method.
In contrast, other methods rapidly deteriorate under noisy data, as they suffer from the domain shift caused by severe noise, which is not observed during the training neural network under the supervised set-up.

Figure~\ref{fig:noise_reduction} shows the visual comparison of reconstructed images of NeRF for inputs with different noise levels against E2VID+~\cite{Stoffregen:2020:SOTA}, which is the state-of-the-art method for intensity image reconstruction.
While E2VID+~\cite{Stoffregen:2020:SOTA} is vulnerable even to small noise, the intrinsic multi-view consistency of Ev-NeRF results in stable performance regardless of noise level and alleviates the effect of noise.
Additional results are provided in the supplementary material.

\begin{figure*}[ht]
\centering
\includegraphics[width=\linewidth]{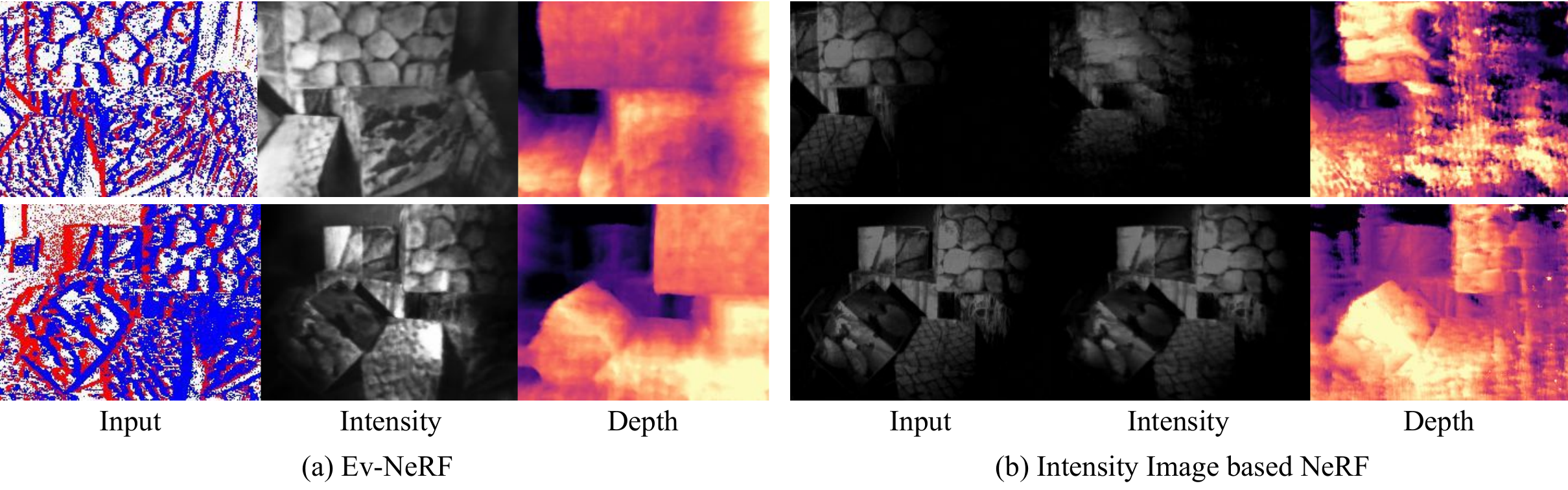}
\caption{Qualitative comparison on intensity image reconstruction and depth estimation of the NeRF volume trained with (a) event data
and (b) intensity images captured in low-light conditions.}
\label{fig:intensity_nerf}
\end{figure*}

\begin{table*}[ht!]

\centering
\resizebox{ \linewidth}{!}{
\begin{tabular}{l|ccc|ccc||ccc|ccc}

\hline

Task & \multicolumn{3}{c|}{E2VID+~\cite{Stoffregen:2020:SOTA} + NeRF} & 
\multicolumn{3}{c||}{Ev-NeRF} & 
\multicolumn{3}{c|}{E2VID+~\cite{Stoffregen:2020:SOTA} + NeRF Noise 0.3} & 
\multicolumn{3}{c}{Ev-NeRF Noise 0.3} \\

\hline
\hline

 & Abs Rel $\downarrow$ & RMSE $\downarrow$  & Sq Rel $\downarrow$ & Abs Rel $\downarrow$ & RMSE $\downarrow$  & Sq Rel $\downarrow$ & Abs Rel $\downarrow$ & RMSE $\downarrow$  & Sq Rel $\downarrow$ & Abs Rel $\downarrow$ & RMSE $\downarrow$  & Sq Rel $\downarrow$ \\
 \hline
Depth & 0.082 & 0.092 & 0.101 & 0.034 & 0.052 & 0.053 & 0.099 & 0.112 & 0.133 & 0.038 & 0.056 & 0.058 \\

\hline
 \hline

 & MSE $\downarrow$ & SSIM $\uparrow$ & LPIPS $\downarrow$ & MSE $\downarrow$ & SSIM $\uparrow$ & LPIPS $\downarrow$ & MSE $\downarrow$ & SSIM $\uparrow$ & LPIPS $\downarrow$ & MSE $\downarrow$ & SSIM $\uparrow$ & LPIPS $\downarrow$ \\
 \hline
Intensity & 0.02 & 0.89 & 0.15 & 0.01 & 0.92 & 0.07 & 0.03 & 0.83 & 0.19 & 0.01 & 0.90 & 0.08   \\
\hline
Novel View & 0.03 & 0.85 & 0.17 & 0.01 & 0.91 & 0.07 & 0.04 & 0.75 & 0.25 & 0.01 & 0.89 & 0.09  \\
\hline
\end{tabular}
}
\caption{~Quantitative results for depth estimation, intensity image reconstrunction, and novel view synthesis on synthetic datasets. Ev-NeRF consistently outperforms the designed baseline even in noisy conditions.}
\vspace{-0.6em}
\label{tab:supp_syunthetic}
\end{table*}
\paragraph{HDR Image Reconstruction}

In addition to noise reduction, under extreme light conditions, the reconstructed images from Ev-NeRF naturally contain high dynamic range information without further processing.
Figure~\ref{fig:intensity_nerf}(a) shows the qualitative results on HDR imaging. 
Compared to the intensity images concurrently captured in low-light set-up ((b), left), which are unknown to the algorithm, the sensor measurements detect the subtle details within the scene ((a), left), which are compiled to produce the HDR intensity images ((a), middle).
This is because Ev-NeRF generates the intensity values to reflect the fine-grained changes in illumination without saturation and is agnostic to any prior absolute values of the intensity that might be clipped to a smaller range.
While we use intensity images to find the poses, the neural volume trained with event data contains variations and details that could not be captured with low-quality intensity images, especially in challenging lighting.

In Figure~\ref{fig:intensity_nerf}(b), we also present the rendering using the neural volume of ordinary NeRF for comparison, which is trained with the intensity images of the same sequence obtained from a DAVIS camera~\cite{daviscamera}.
The intensity-based NeRF is trained with the MSE loss between the rendered and measured intensity, following~\cite{mildenhall:2020:nerf}.
The reconstructed images and depths are superior when using Ev-NeRF, therefore fully exploiting the dynamic range of the sensor.

Under extreme lighting, not only do intensity frames suffer from spurious noise, but also the event data is mixed with severe noise, and the performance of vision-based approaches severely deteriorates~\cite{Kim_2021_ICCV,ev-tta}.
The multi-view consistency in Ev-NeRF compensates for the uncharacteristic noises and reliably reconstructs novel views and depth information beyond the level obtained from vanilla NeRF using intensity images.

\subsection{Scene Structure Estimation}
\label{sec:result_structure}

\paragraph{Evaluation with Synthetic Data}

\begin{figure}[h]
\vspace{-0.5em}
\centering
\includegraphics[width=0.83\linewidth]{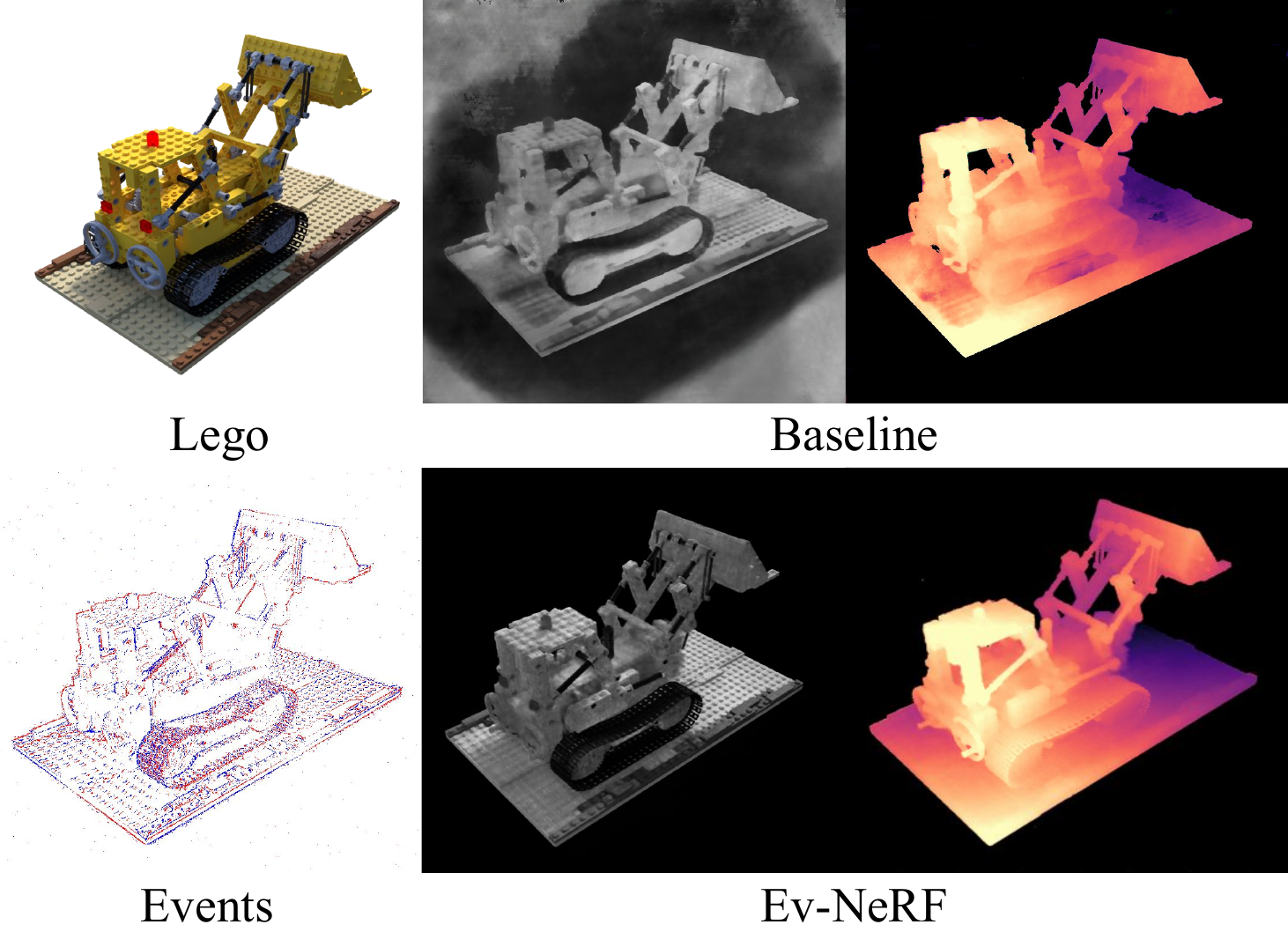}
\caption{
Qualitative results for intensity image, dense depth by baseline (E2VID+~\cite{Stoffregen:2020:SOTA} + NeRF) and Ev-NeRF.}
\label{fig:fig1}
\vspace{-1.2em}
\end{figure}

The NeRF volume aggregates multiple measurements into a coherent structure such that we can generate images from arbitrary viewpoint. 
Also, the density values of NeRF volume can provide estimates for 3D structure as in Equation~\ref{eq:nerf_depth}.
Since the real-world datasets for event-based vision lack dense depth annotations, the synthetic data can provide quantitative evaluation on the created NeRF volume.
Also, the 3D points found with existing event-based algorithms are limited to sparse reconstruction that can generate events and no prior work is able to reconstruct novel view images.
Therefore, we design baseline for quantitatively comparison.

For the synthetic data, we capture a high frame rate video from typical trajectories for NeRF with Blender, and the event simulator~\cite{v2e} generates events with realistic noise from the video.
For the baseline, we first reconstruct intensity images from events using E2VID+~\cite{Stoffregen:2020:SOTA} and then use them to learn vanilla NeRF~\cite{mildenhall:2020:nerf}.
The proposed method serves as a powerful baseline that can reconstruct dense depth and an intensity image from a novel view.
Figure~\ref{fig:fig1} shows an object in Blender, generated events, and the pairs of an intensity image and the depth generated from the designed baseline and Ev-NeRF.

Table~\ref{tab:supp_syunthetic} compares the depth estimation results using three metrics:
average relative error (Abs Rel), squared relative difference (Sq Rel), root mean squared error (RMSE).
Also the reconstructed intensity images in the given or novel views against the ground truth intensity images are compared using three metrics: mean squared error (MSE), structural similarity (SSIM) and perceptual similarity (LPIPS)~\cite{zhang2018perceptual}.
In baseline, images reconstructed through E2VID+~\cite{Stoffregen:2020:SOTA} lack multi-view consistency and result in an artifact, which reduces the accuracy of 3D structures. 
Ev-NeRF directly integrates events and Table~\ref{tab:supp_syunthetic} indicates that it is superior to the proposed baseline for depth reconstruction.
Also, the high-quality 3D structure aids in more accurate reconstruction of intensity images or novel-view synthesis, especially in the presence of noise.

\paragraph{Evaluation with Real Data}

\begin{figure}[t]
\centering
\includegraphics[width=0.9\linewidth]{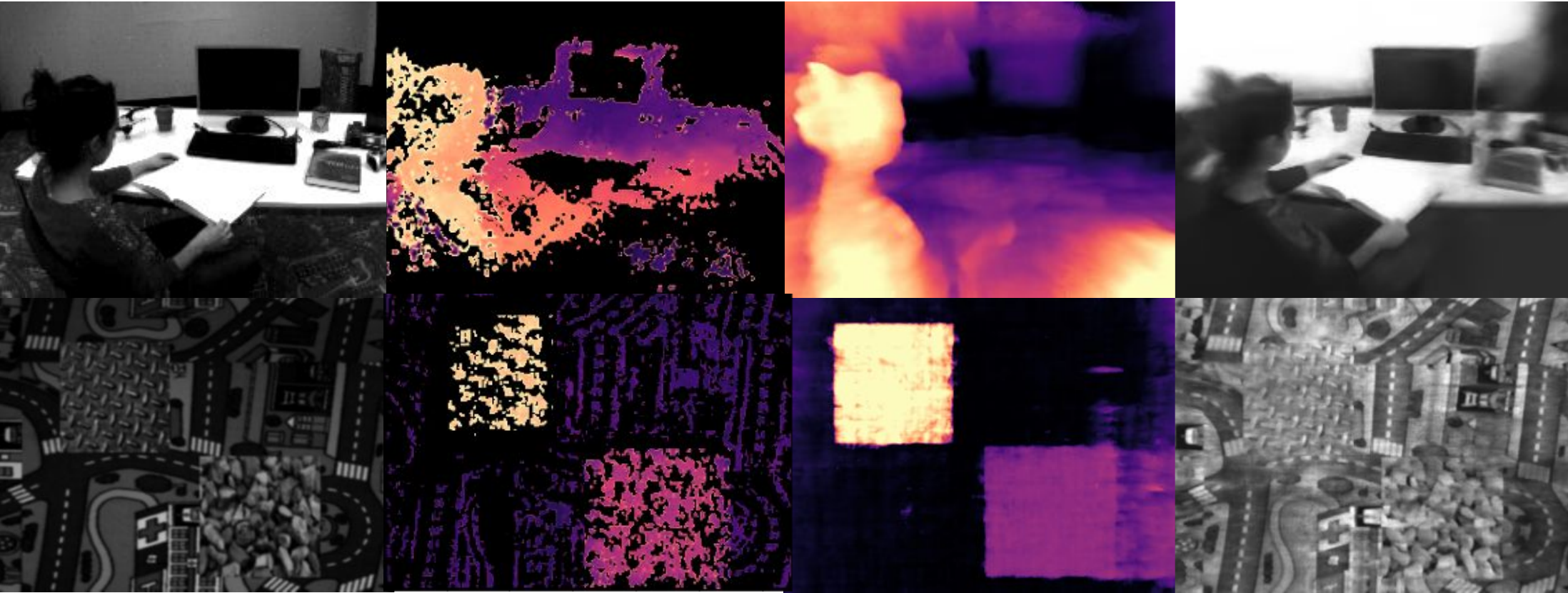}
\caption{Qualitative comparison on 3D structure estimation.
Ground-truth intensity frame, semi-dense depth by~\cite{zhou:2018:3d}, dense depth, and intensity image by Ev-NeRF in order.}
\label{fig:slam}
\vspace{-0.7em}
\end{figure}

\begin{figure}[t]
\centering
\includegraphics[width=0.93\linewidth]{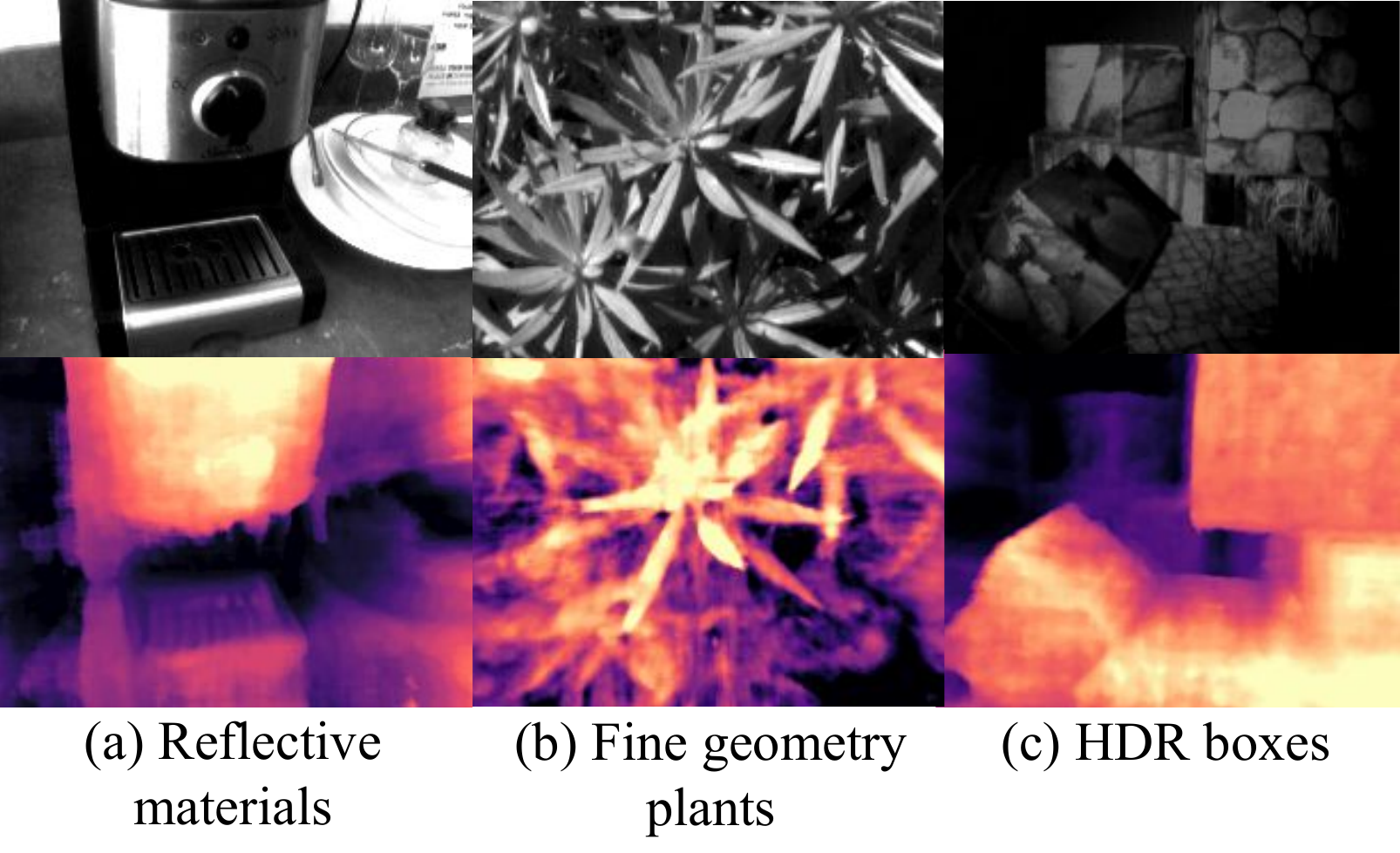}
\caption{Qualitative results on depth estimation on a real dataset. The intensity images (top) are not observed by the algorithm.}
\label{fig:depth_qualititive}
\vspace{-1.0em}
\end{figure}

There is no prior work that is bound to the exact set-up as Ev-NeRF, and no real-world dataset with ground-truth is available to quantitatively evaluate the dense 3D structure.
As an alternative, we compare the acquired geometry against the 3D information extracted from an event-based SLAM approach~\cite{zhou:2018:3d}.
Event-based SLAM utilizes measurements from additional hardware  and compute the camera pose and the geometric structure.
Specifically, Zhou et al.~\cite{zhou:2018:3d} use a pair of temporally synchronized event cameras.
On the other hand, Ev-NeRF uses a single event camera with known poses.
Even though the detailed setup is different, both estimate 3D structure from the measurements of a moving sensor observing a static scene, whose results are provided in Figure~\ref{fig:slam}.
Zhou et al.~\cite{zhou:2018:3d} reconstruct depth only in the area where the events have occurred, and therefore recover semi-dense 3D structure (second column in Figure~\ref{fig:slam}).
In contrast, Ev-NeRF reconstructs implicit 3d volume and can produce denser depth with intensity images (third and last columns in Figure~\ref{fig:slam}).

We can also find qualitative results from the dataset used in Sec.~\ref{sec:result_robust}.
Figure~\ref{fig:intensity_nerf} contains the depth estimation from low lighting conditions.
Additionally, Figure~\ref{fig:depth_qualititive} shows that Ev-NeRF creates reasonable depth estimates for challenging real-world scenes with reflective materials, fine geometry details, or HDR measurements where existing approaches might fail.

\begin{table}[t]

\centering
\resizebox{\linewidth}{!}{
\begin{tabular}{l|cc|cc|cc}

\hline

& \multicolumn{2}{c|}{MSE $\downarrow$} & 
\multicolumn{2}{c|}{SSIM $\uparrow$} & 
\multicolumn{2}{c}{LPIPS $\downarrow$}\\

\hline

Scene & Given & Novel  & Given & Novel & Given & Novel \\

\hline

office\_zigzag & 0.03 & 0.04 & 0.42 & 0.41 & 0.27 & 0.28    \\
office\_spiral & 0.03 & 0.03 & 0.41 & 0.40 & 0.27 & 0.27   \\
boxes & 0.04 & 0.04 & 0.48 & 0.46 & 0.31 & 0.33    \\
dynamic\_6dof & 0.19 & 0.20 & 0.26 & 0.26  & 0.41 & 0.43   \\

\hline

reflective\_materials & 0.05 & 0.06 & 0.40 & 0.40 & 0.35 & 0.38   \\
high\_texture\_plants & 0.03 & 0.03 & 0.44 & 0.42 & 0.34 & 0.36   \\
still\_life & 0.03 & 0.04 & 0.53 & 0.52 & 0.18 & 0.18   \\
\hline

\end{tabular}
}

\caption{Quantitative results of novel view reconstruction. 
Compared to given views provided during training, only a small performance gap is observed to reconstruct images of novel views.}

\label{tab:novelview}
\vspace{-0.8em}
\end{table}

We also evaluate the performance of novel view synthesis.
We divide each sub-sequence in the dataset into a training and test set, and synthesize the images at the viewpoints included in the test set with Ev-NeRF trained with the training set.
Table~\ref{tab:novelview} compares the reconstructed images against ground truth intensity images.
The reported metrics for novel-view images are compatible with the results for the given view.
Thus, the ability of NeRF is nicely transferred to Ev-NeRF.
\section{Conclusions and Future Work}

We present Ev-NeRF, which learns the implicit volume of the neural radiance field from the raw stream of events generated by a neuromorphic camera.
The inherent multi-view consistency creates a representation remarkably robust to noisy inputs, which is a critical challenge for using a neuromorphic sensor, yet exploits the subtle brightness changes detected from the sensor. 
Further, the created NeRF volume can generate intensity images or estimate depth, whose quality is comparable to many existing supervised methods exclusively designed to solve a specific task.
To the best of our knowledge, Ev-NeRF is the first attempt to incorporate the NeRF formulation with raw event data and can advance with the abundant subsequent works that overcome the limitations of NeRF, such as handling dynamic scenes to better incorporate the fast temporal resolution of the sensor~\cite{dynamic1,dynamic2}, reducing training and rendering time~\cite{mueller2022instant,sun2021direct,yu2021plenoxels}, and alleviating the requirements of known camera poses~\cite{wang2021nerfmm,lin2021barf}.

\paragraph{Acknowledgements} This research was supported by the National Research Foundation of Korea (NRF) grant funded by the Korea government(MSIT) (No. 2020R1C1C1008195), Samsung Electronics Co., Ltd, and Creative-Pioneering Researchers Program through Seoul National University. Inwoo Hwang is grateful for financial support from Hyundai Motor Chung Mong-Koo Foundation. 

\clearpage

{\small
\bibliographystyle{ieee_fullname}
\bibliography{ev_nerf}
}

\appendix

\section{Additional Implementation Details}

\begin{figure}[h]
\centering
\includegraphics[width=0.8\linewidth]{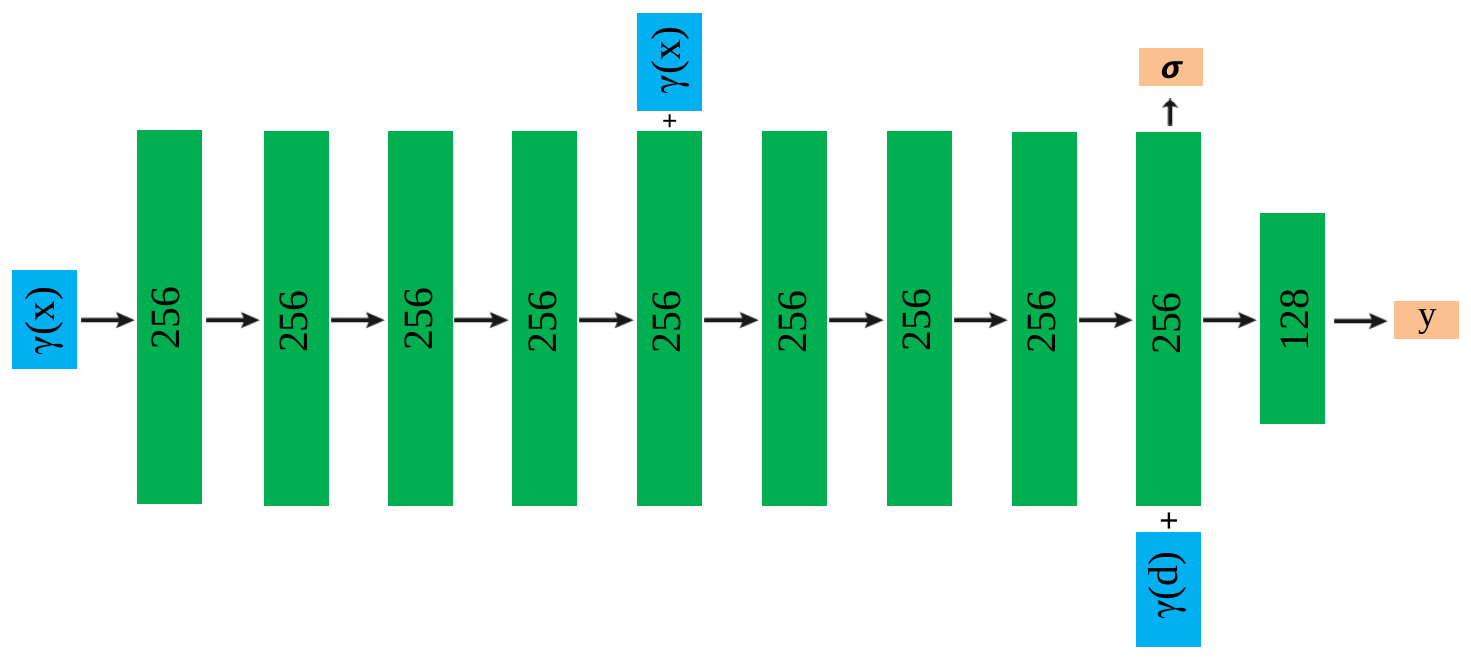}
\caption{Network Architecture of Ev-NeRF. 
The network consists of fully-connected layers. 
The numbers in the colored blocks indicate the dimension of the corresponding layers.}
\label{fig:supp_figures_archi}
\end{figure}

We implement Ev-NeRF using PyTorch using the NeRF formulation.
For positional encoding, we use 10 frequencies for $\mathbf{x}$ and 4 for $\mathbf{d}$.
The weight $\lambda$ of $\mathcal{L}_\text{thres}$ in Equation 8 is set large to 1000 to avoid thresholds from continuing to decrease.
For all experiments, we use the Adam optimizer~\cite{Kingma:2015:Adam} with a learning rate of  $5\times 10^{-4}$.
Ev-NeRF takes about an hour in RTX 3090 GPU to train per scene and 1.5 seconds to render a single image.

Similar to NeRF, the neural network takes the 3D coordinate and the ray direction as input and outputs the volume density and emitted radiance.
Sinusoidal positional encoding, $\gamma \left( \cdot \right)$, is applied to input variables.
Figure~\ref{fig:supp_figures_archi} shows the detailed architecture of our network.
The architecture mostly follows that of NeRF~\cite{mildenhall:2020:nerf}, however it predicts the emitted luminance value instead of color values.
Following NeRF~\cite{mildenhall:2020:nerf}, we add zero-mean, unit-variance Gaussian random noise to the density $\sigma$ for slightly improved performance.

\section{Dataset Description}

For the real-world data, we use a sub-sequence of the event sequences from
IJRR~\cite{Mueggler:2017:dataset}, HQF~\cite{Stoffregen:2020:SOTA} and Stereo DAVIS~\cite{zhou:2018:3d} for training. 
The data includes intensity images at regular time intervals (about 24 Hz) and asynchronous event data.
These datasets are generated with a  DAVIS240C~\cite{daviscamera} event camera and both intensity images and events have a resolution of 240$\times$180.
In our setup, we assume camera poses are given, which are calculated from running SfM~\cite{schoenberger:2016:colmap-sfm,schoenberger:2016:colmap-mvs} with the intensity frames.
Except for this process, the intensity frames are not available during training and are used only for evaluation. 
We use sub-sequences with a length of 50 to 100 intensity frames for training and the exact frame index corresponding to the original dataset~\cite{Mueggler:2017:dataset,Stoffregen:2020:SOTA} is described in Table~\ref{tab:supp1}.
For comparison with event-based SLAM, we use two sequences ({simulation\_3planes, reader}) from~\cite{zhou:2018:3d}.

\begin{figure}[h]
\centering

\includegraphics[width=\linewidth]{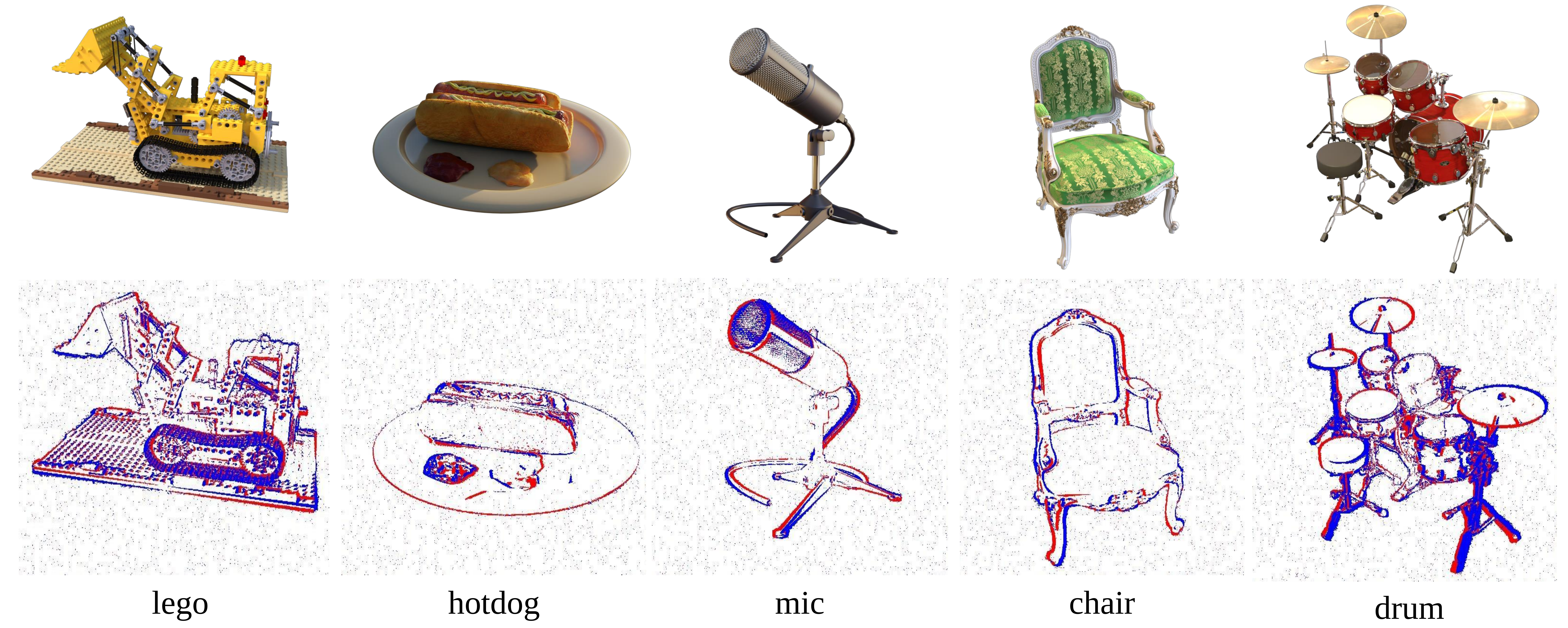}
\caption{Visualization of generated events for each scene.}
\label{fig:figures_supp3}
\end{figure}

For the synthetic data, we examine the extracted scene structure using models widely used for NeRF~\cite{mildenhall:2020:nerf}, namely lego, hotdog, mic, drums, and chair.
Figure~\ref{fig:figures_supp3} shows examples of models and generated events.

\begin{table}[h]

\centering
\resizebox{\linewidth}{!}{
\begin{tabular}{c|c|cc}
\hline

Dataset & Scene & start frame index & end frame index  \\ 
\hline
\multirow{5}{*}{IJRR} &
office\_zigzag & 0 & 100    \\
& office\_spiral & 0 & 100   \\
& boxes& 230 & 330   \\
& dynamic\_6dof & 30 & 130   \\
& hdr\_boxes & 70 & 120   \\
\hline
\multirow{3}{*}{HQF} &
reflective\_materials & 60 & 150  \\
& high\_texture\_plants & 930 & 1000   \\
& still\_life & 300 & 350  \\
\hline
\multirow{2}{*}{Stereo DAVIS} &
monitor & 0 & 100  \\
& reader & 0 & 100   \\
\hline
\end{tabular}
}
\caption{
The start and end frame index for intensity frames used for our experiments.
We use sub-sequences of original datasets, namely IJRR~\cite{Mueggler:2017:dataset}, HQF~\cite{Stoffregen:2020:SOTA} and Stereo DAVIS~\cite{zhou:2018:3d}.}

\label{tab:supp1}
\end{table}

\section{Description on Convergence}

\begin{figure}[ht]
\centering
\includegraphics[width=0.95\linewidth]{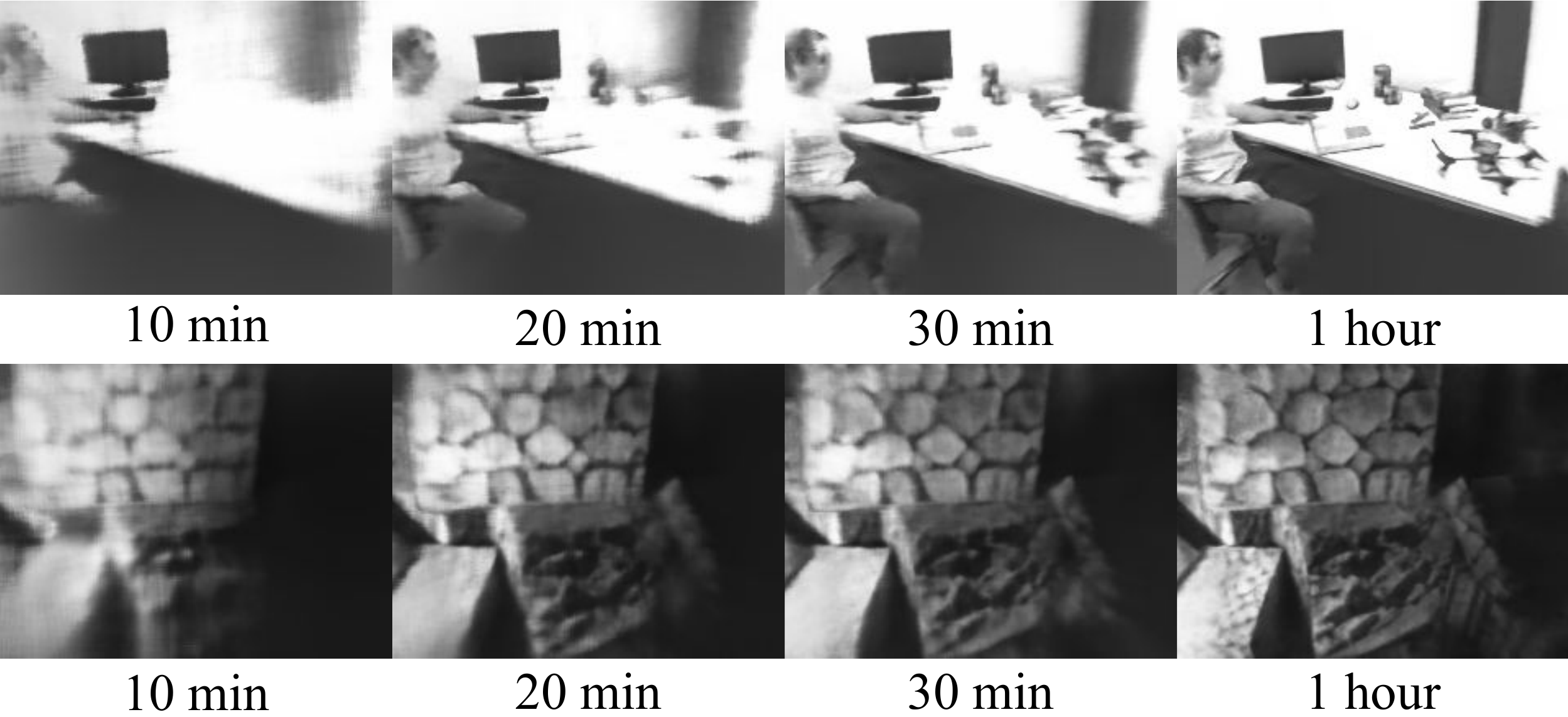}
\caption{Intermediate results over the training time for each scene.}
\label{fig:figures_time}
\end{figure}

\begin{table*}[ht!]

\begin{center}

\resizebox{0.9\linewidth}{!}{
\begin{tabular}{l|ccc|ccc|ccc}

\hline

& \multicolumn{3}{c|}{MSE $\downarrow$} & 
\multicolumn{3}{c|}{SSIM $\uparrow$} & 
\multicolumn{3}{c}{LPIPS $\downarrow$}\\

\hline

Scene & w/o joint & w/o noise inj. & full  & w/o joint & w/o noise inj. & full & w/o joint & w/o noise inj. & full \\ \hline
office\_zigzag & 0.03 & 0.04  & \textbf{0.03} & 0.40 & 0.41 & \textbf{0.42} & 0.29 & 0.27 & \textbf{0.27}    \\
office\_spiral & 0.04 & 0.04 & \textbf{0.03} & 0.41 & 0.41 & \textbf{0.41} & 0.29 & 0.28 & \textbf{0.27}   \\
boxes & 0.04 & 0.06 & \textbf{0.04} & 0.47 & 0.45 & \textbf{0.48} & 0.32 & 0.33 & \textbf{0.31}    \\
dynamic\_6dof & 0.21 & 0.24 & \textbf{0.19} & 0.25 & 0.24 & \textbf{0.26} & 0.42 & 0.43 & \textbf{0.41}   \\

\hline
reflective\_materials & 0.06 & 0.07 & \textbf{0.05} & 0.39 & 0.38 & \textbf{0.40} & 0.35 & 0.36 & \textbf{0.35}   \\
high\_texture\_plants & 0.04 & 0.03 & \textbf{0.03} & 0.43 & 0.43 & \textbf{0.44} & 0.34 & 0.35 & \textbf{0.34}   \\
still\_life & 0.04 & 0.05 & \textbf{0.03} & 0.52 & 0.51 & \textbf{0.53} & 0.21 & 0.19 & \textbf{0.18}   \\
\hline

monitor & 0.05 & 0.08 & \textbf{0.03} & 0.30 & 0.26 & \textbf{0.32} & 0.38 & 0.39 & \textbf{0.37}   \\
reader & 0.11 & 0.12 & \textbf{0.09} & 0.44 & 0.43 & \textbf{0.45} & 0.38 & 0.36 & \textbf{0.35}   \\
\hline

\end{tabular}
}
\caption{Ablation study on the effect of joint training of the sensor threshold values and noise injection. 
The columns display results without joint training, without noise injection, and the full model, respectively.
Our full training method shows the optimal reconstructed image quality.}
\label{tab:ablation2}
\end{center}
\end{table*}

While events provide only changes of brightness, we empirically found it converges to reliable absolute brightness as the learning progresses.
The supplementary video shows how Ev-NeRF converges on absolute brightness as training progresses.
Figure~\ref{fig:figures_time} contains a few representative images.
Ev-NeRF obtains a rough 3D structure in about 10 min, and the subsequent timesteps focus on capturing further details.

\section{Ablation Studies}

\paragraph{Joint Training} 
As shown in Equation (8), we propose joint training that concurrently optimizes $\Delta I$ and the threshold values $B_j^+, B_j^-$ of all timestamps with $\lambda = 1000$.
We verify the advantage of our joint training on intensity reconstruction in Table~\ref{tab:ablation2}.
We compare the proposed joint training against the ablated version with pre-fixed threshold values $\pm 0.3$ and $\lambda=0$.
Results show that the joint training scheme is beneficial to the reconstructed image quality.

\begin{figure}[t]
\centering
\includegraphics[width=0.9\linewidth]{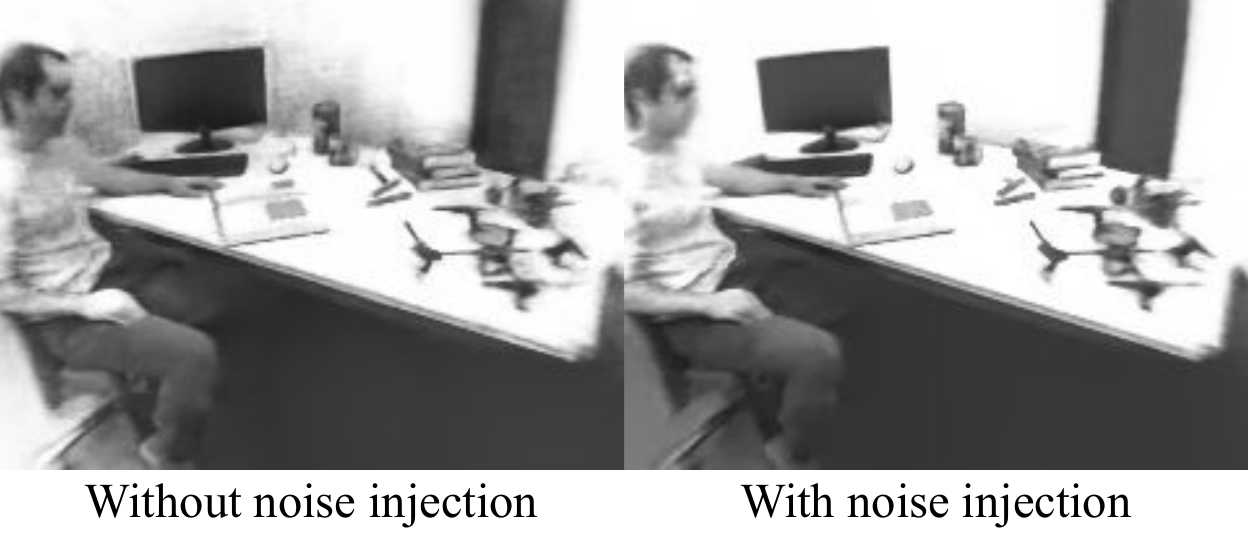}
\caption{Visual analysis on noise injection, which helps solve ambiguity in places such as walls where events do not occur. }
\label{fig:noise_reg}
\end{figure}

\paragraph{Noise Injection}
We verify that the additional random noise slightly improves the quality of Ev-NeRF.
When we compile the training data of events occurred during time slice $[T_j , T_{j+1})$,  we add random events whose amount is 5\% of the number of events that occurred at time. 
Table~\ref{tab:ablation2} numerically compares the quality of reconstructed images with random noise injection against the original event slice $\mathcal{E}_j$ and verifies that the noise injection is beneficial to the image quality.
Figure~\ref{fig:noise_reg} visualizes the effect of noise injection.
The additional noise helps the neural volume to resolve ambiguity in the areas where events do not occur, such as the solid-colored wall behind the monitor.

\section{Additional Results}
Here we show more examples of quantitative and qualitative results. 
The results are presented in various scenes with three baselines:
E2VID~\cite{Rebecq:2019cvpr:intensity}, E2VID+~\cite{Stoffregen:2020:SOTA} and ssl-E2VID~\cite{paredes:2021:SSL-intensity-baseline},
which are designed to reconstruct intensity images.

\subsection{Intensity Image Reconstruction}
\paragraph{Additional Quantitative Results}

\begin{table*}[t]

\centering
\resizebox{\linewidth}{!}{
\begin{tabular}{l|cccc|cccc|cccc}

\hline 

& \multicolumn{4}{c|}{MSE $\downarrow$} & 
\multicolumn{4}{c|}{SSIM $\uparrow$} & 
\multicolumn{4}{c}{LPIPS $\downarrow$}\\

\hline 

Scene & E2VID & E2VID+ & ssl-E2VID & Ours & E2VID & E2VID+ & ssl-E2VID & Ours & E2VID & E2VID+ & ssl-E2VID & Ours \\

\hline

office\_zigzag & 0.07 & \underline{0.05} & 0.08 & \textbf{0.03} & 0.38 & \underline{0.39} & 0.34 & \textbf{0.42} & 0.34 & \textbf{0.25}  & 0.40 & \underline{0.27}    \\
office\_spiral & 0.06 & \underline{0.05} & 0.07 & \textbf{0.03} & 0.38 & \underline{0.39} & 0.37 & \textbf{0.41} & 0.35 & \underline{0.27} & 0.39 & \textbf{0.27}    \\
boxes & 0.06 & \textbf{0.03} & 0.08 & \underline{0.04} & 0.47 & \textbf{0.60} & 0.45 & \underline{0.48} & 0.31 & \textbf{0.20}  & 0.37 & \underline{0.31}  \\
dynamic\_6dof & \underline{0.12} & \textbf{0.06} & 0.15 & 0.19 & \underline{0.29} & \textbf{0.34} & 0.28 & 0.26 & \underline{0.40} & \textbf{0.33}  & 0.54 & 0.41     \\

\hline 

reflective\_materials & 0.07 & \underline{0.05} & 0.08 & \textbf{0.05} & 0.39 & \textbf{0.45} & 0.30 & \underline{0.40} & \underline{0.31} & \textbf{0.24} & 0.38 &  0.35  \\
high\_texture\_plants & 0.04 & \textbf{0.02} & 0.05 & \underline{0.03} & 0.42 & \textbf{0.55} & 0.42 & \underline{0.44} & \underline{0.21} & \textbf{0.12} & 0.23 & 0.34    \\
still\_life & 0.05 & \textbf{0.02} & 0.08 & \underline{0.03} & 0.50 & \textbf{0.61} & 0.40 & \underline{0.53} & 0.23 & \textbf{0.13} & 0.29 & \underline{0.18}  \\

\hline

monitor & 0.04 & \underline{0.04} & 0.09 & \textbf{0.03} & 0.31 & \textbf{0.36} & \underline{0.33} & 0.32 & 0.39 & \textbf{0.20} & \underline{0.34} & 0.37  \\
reader & \underline{0.07} & \textbf{0.04} & 0.09 & 0.09 & 0.42 & \underline{0.43} & 0.38 & \textbf{0.45} & \underline{0.31} & \textbf{0.25} & 0.40 & 0.35  \\

\hline
\end{tabular}
}
\caption{Quantitative comparison of image reconstruction on scenes from the IJRR~\cite{Mueggler:2017:dataset}, HQF~\cite{Stoffregen:2020:SOTA} and Stereo DAVIS~\cite{zhou:2018:3d} dataset. The results with the best performance are in bold. We additionally underline the runner-up metric.}
\label{tab:intensity}

\end{table*}

Table~\ref{tab:intensity} displays the quantitative comparison against baseline methods with real-world datasets, namely IJRR~\cite{Mueggler:2017:dataset},  HQF~\cite{Stoffregen:2020:SOTA}, and Stereo DAVIS dataset~\cite{zhou:2018:3d}.
We compared using three metrics: mean squared error (MSE), structural similarity (SSIM) and perceptual similarity (LPIPS)~\cite{zhang2018perceptual}.
For real-world data, Ev-NeRF outperforms ssl-E2VID and is on par with E2VID and  E2VID+ without observing the ground truth intensity frames.

However, we used sub-sequences of real datasets and event camera trajectory from sub-sequences is not optimal for NeRF, which is typically trained with cameras located on the hemisphere around the object from roughly constant distances.
This causes Ev-NeRF to show lower performance compared to E2VID+ for some sequences.
As mentioned in Table 1 from the main paper, with typical trajectories for NeRF, the quality of intensity images of Ev-NeRF is then consistently superior to the baseline, especially in the presence of noise.

\paragraph{Additional Qualitative Results for Intensity Image and Novel View Synthesis}

Figure~\ref{fig:figures_supp2} contains the results of intensity image reconstruction for all of the sequences we use dwith a qualitative comparison with various baselines and Figure~\ref{fig:figures_supp5} shows image reconstruction results observed from various camera poses.
Also, the supplementary video contains the event sequences used for training, paired with corresponding reconstructed intensity images and depth results. 
Figure~\ref{fig:figures_supp7} shows novel view synthesis observed from the viewpoints that are not available in the input dataset.
Ev-NeRF maintains comparable performance for all scenes.
Additional results on novel view synthesis are shown in the supplementary video.

\begin{figure}[t]
\centering
\includegraphics[width=0.9\linewidth]{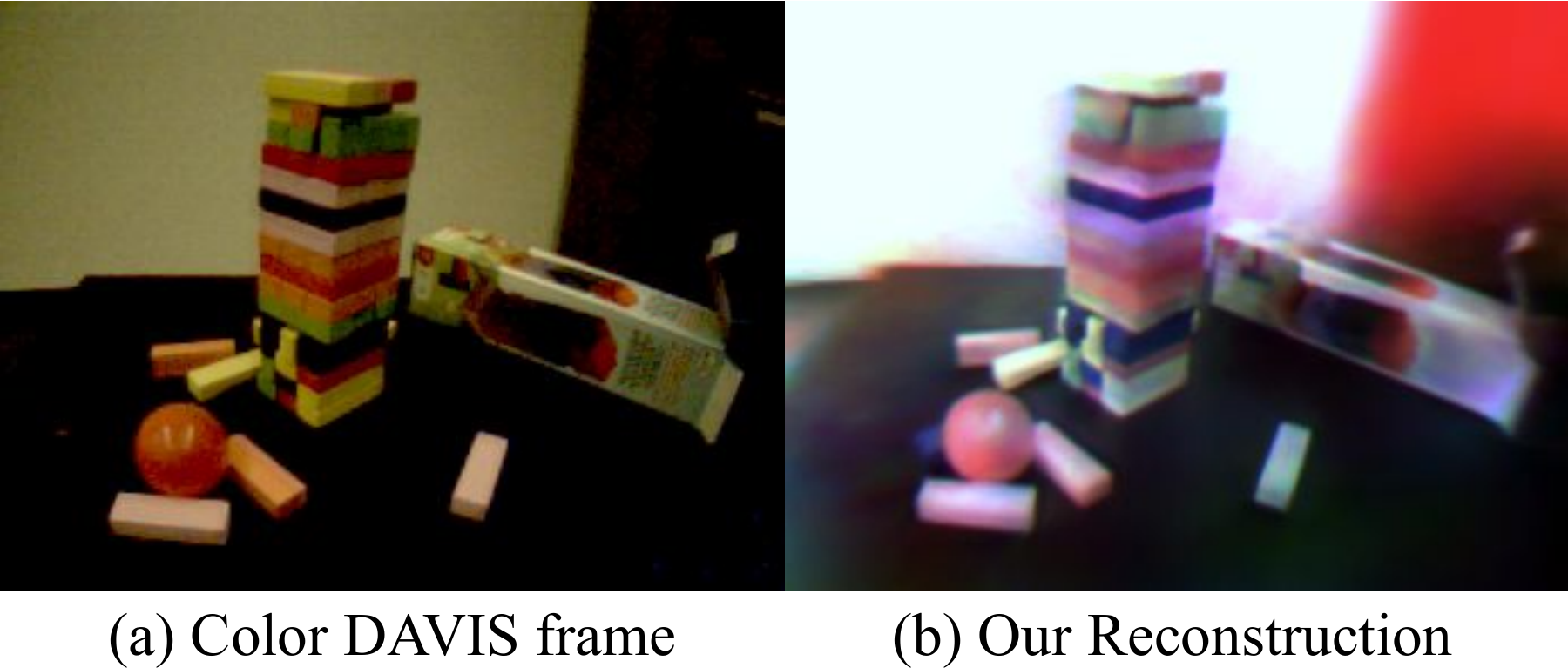}
\caption{Color image reconstruction through Ev-NeRF. }
\label{fig:ced}
\end{figure}

\paragraph{Intensity Image Reconstruction on Different Sensor Resolution}
We further validate the domain-invariance of Ev-NeRF with intensity image reconstruction results on the Color Event Camera Dataset (CED)~\cite{colordataset}.
The data is composed of three channels of color events in a different resolutions (346$\times$240).
All of the datasets presented in other sections are processed in the resolution of 240$\times$180 bound to the sensor resolution.
Ev-NeRF does not assume any fixed resolution of the scene and can be applied in multiple color channels without fine-tuning.
We extend our approach to three color channels following the method suggested in ~\cite{colordataset}; we first reconstruct the intensity images via red, green, and blue channels respectively, and upsample the individual channels of the image to the original resolution to produce a single color image.
Figure~\ref{fig:ced} shows an exemplar color DAVIS frame and reconstructed color image for simple\_jenga scene from CED~\cite{colordataset}.
Ev-NeRF can find the scene structure within reasonable ranges.
More importantly, there is no additional training to account for the domain shift in sensor characteristics or resolution.

\subsection{Noise Resistant Image Reconstruction}
Ev-NeRF is extremely robust under noise and maintains the quality of the reconstruction under data of various noise levels.
In addition to the visual example shown in Figure 4 of the main text, we show results given data corrupted with different levels of noise in Figure~\ref{fig:figures_supp6}.
Figure~\ref{fig:figures_supp4} contains more results of image reconstruction with severe noise of ratio 0.9.
Ev-NeRF exhibits little degradation in performance and therefore can be useful in an extreme environments subject to unknown noise characteristics.

\begin{figure*}[h]
\centering
\includegraphics[width=\linewidth]{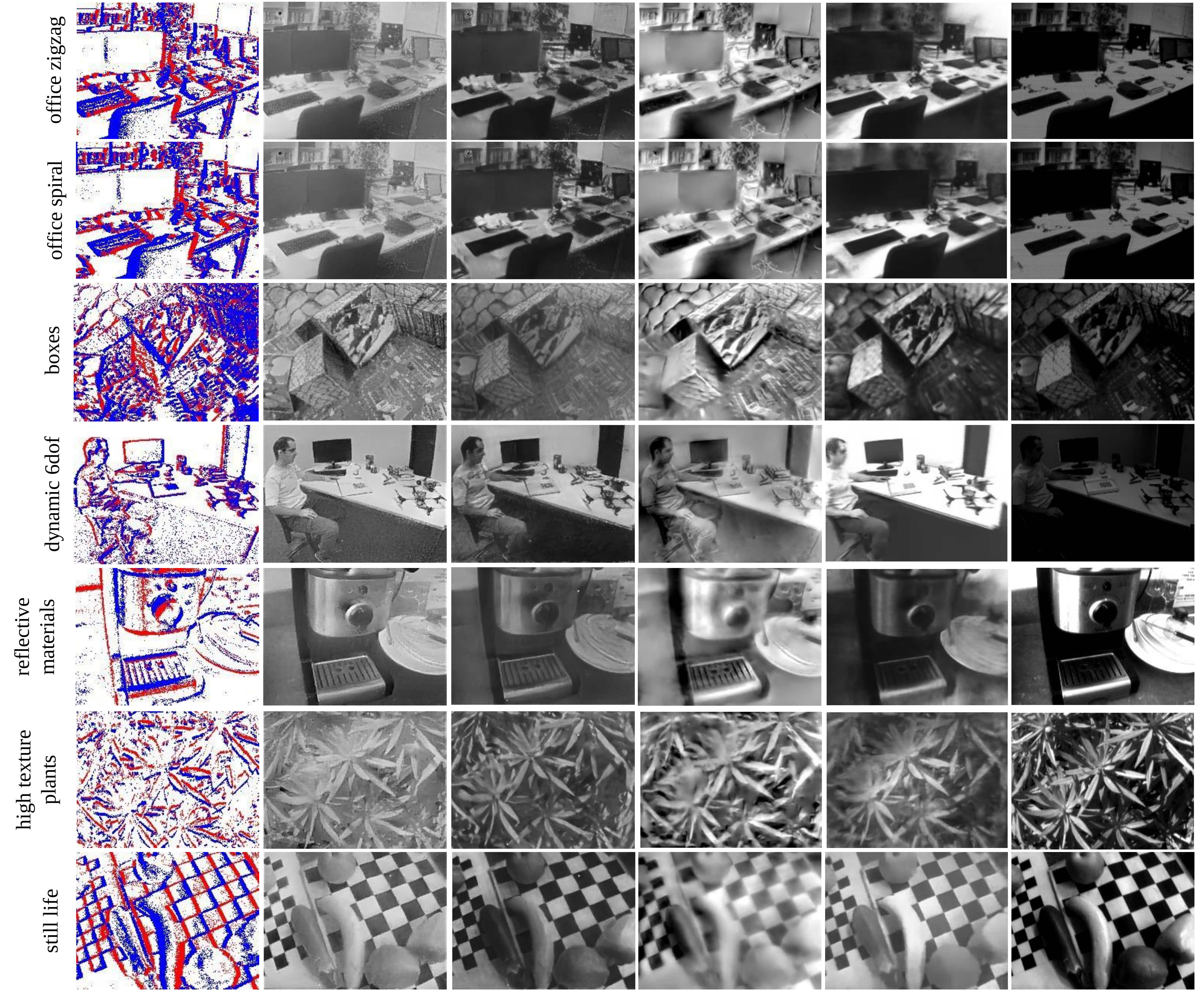}
\vspace{-10.7em}
\end{figure*}

\begin{figure*}[h]
\centering
\includegraphics[width=1.006\linewidth]{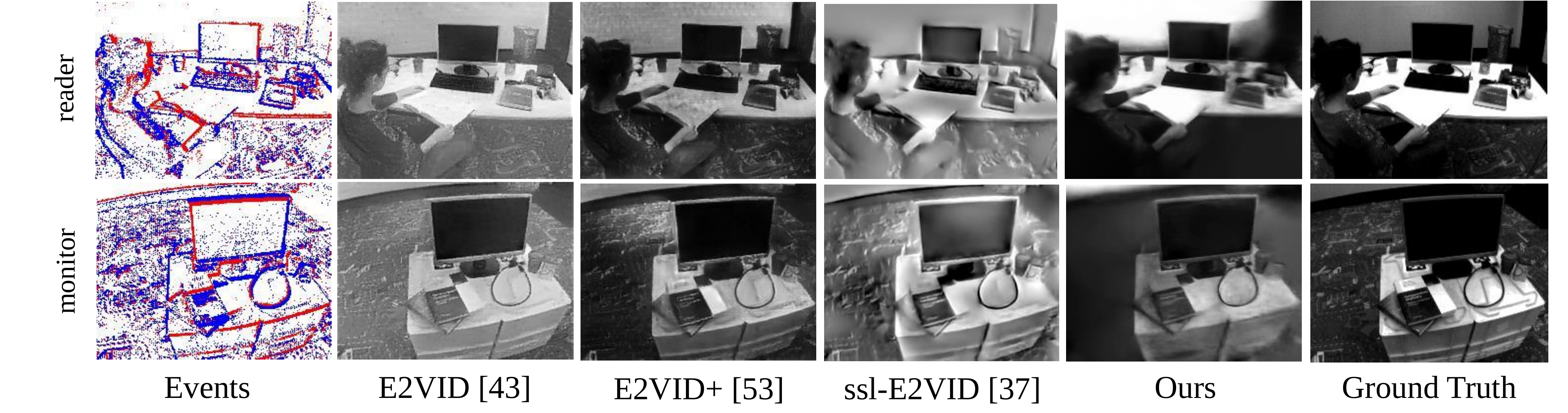}
\caption{Qualitative comparison on intensity image reconstruction under various scenes.}
\label{fig:figures_supp2}
\end{figure*}

\begin{figure*}[th!]
\centering
\includegraphics[width=0.59\linewidth]{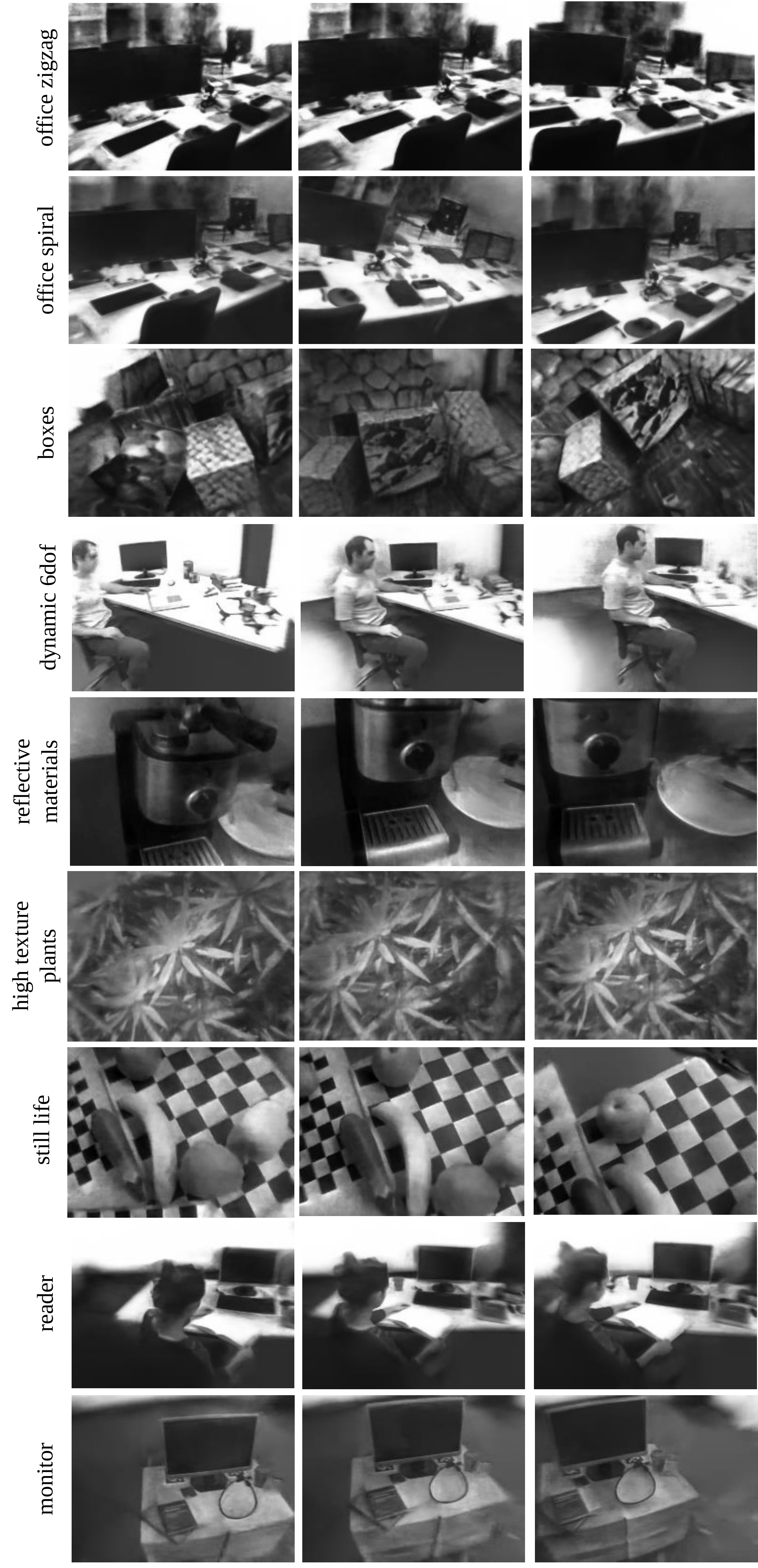}
\caption{Qualitative intensity image reconstruction results at various time steps.}
\label{fig:figures_supp5}
\end{figure*}

\begin{figure*}[h!]
\centering
\includegraphics[width=0.8\linewidth]{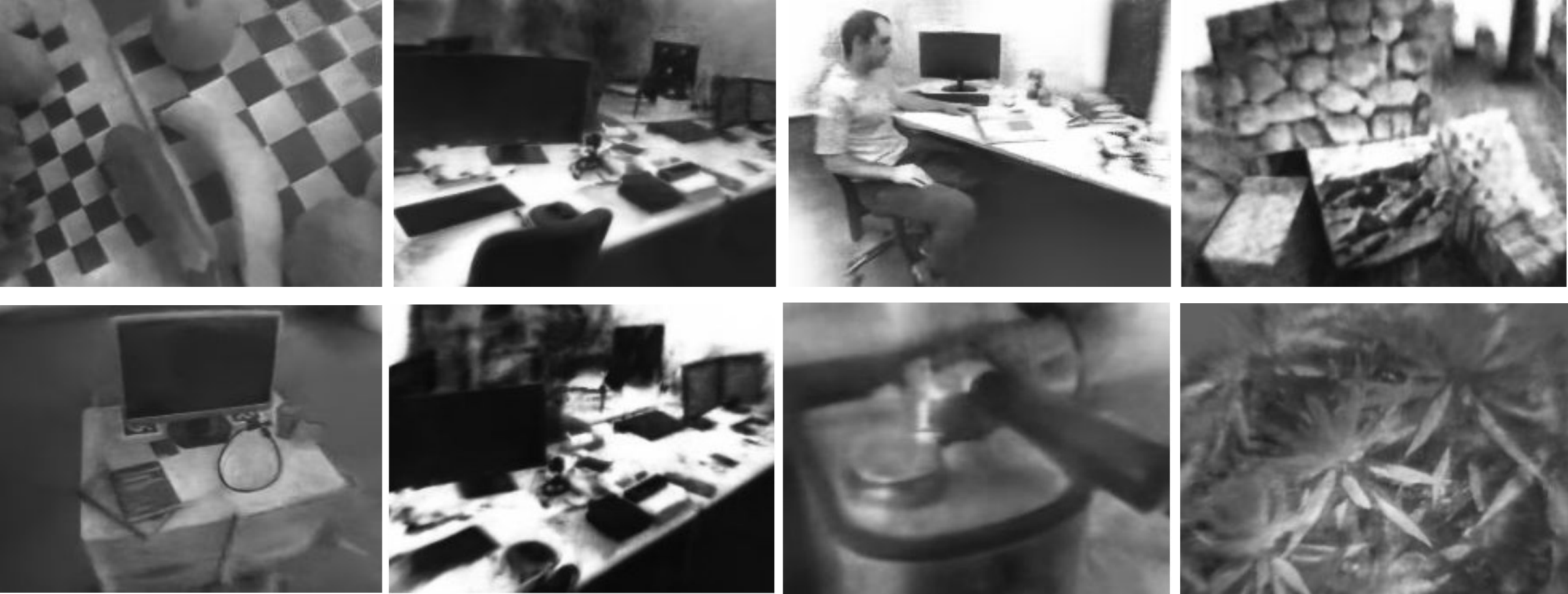}
\caption{Qualitative results for novel view image reconstruction.}
\label{fig:figures_supp7}
\end{figure*}

\begin{figure*}[th]
\centering
\includegraphics[width=0.9\linewidth]{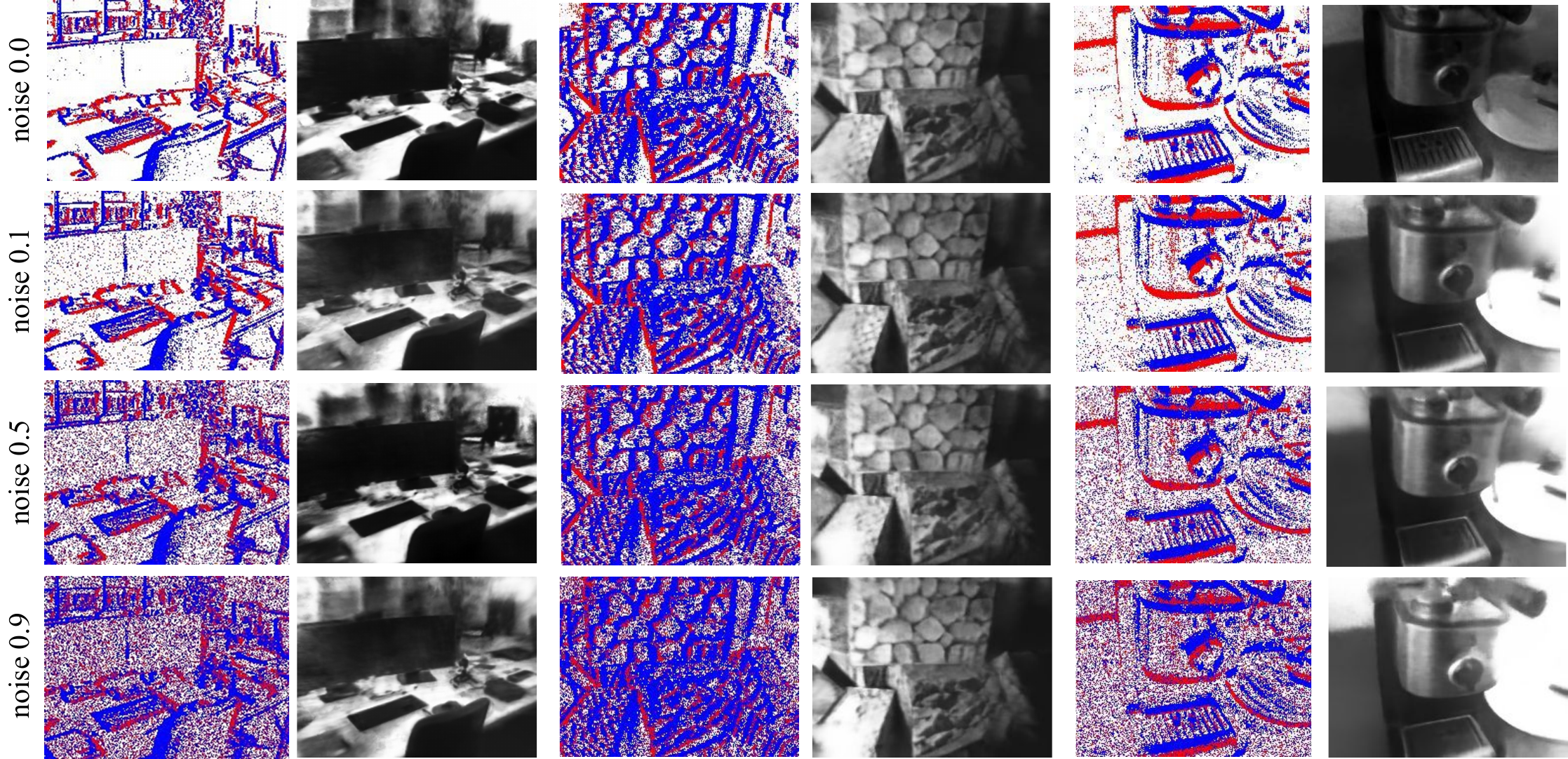}
\caption{Qualitative comparison under different noise levels.}
\label{fig:figures_supp6}
\end{figure*}

\begin{figure*}[th]
\centering
\includegraphics[width=0.7\linewidth]{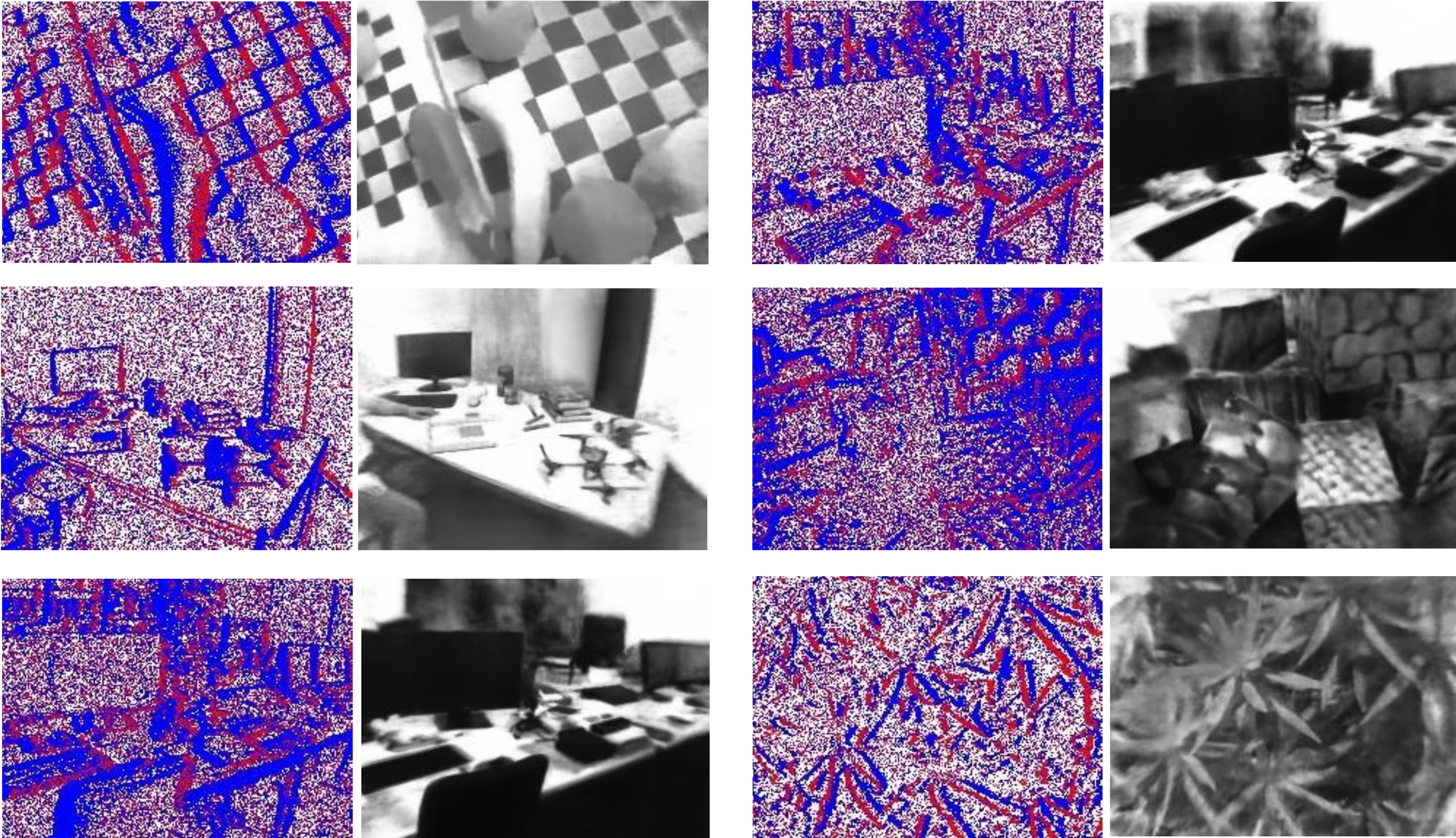}
\caption{Noise reduction of events over various scenes with extreme noise ratio of 0.9.}
\label{fig:figures_supp4}
\end{figure*}

\end{document}